\RequirePackage{fix-cm}
\documentclass[twocolumn]{svjour3}          
\smartqed  
\usepackage{graphicx}
\usepackage{natbib}
\usepackage{multirow}
\usepackage{amsmath}
\usepackage{amssymb}
\usepackage{bbm}
\usepackage{bm}
\usepackage{booktabs}
\usepackage{array} 
\usepackage{blindtext}
\usepackage{comment}
\usepackage{threeparttable}
\usepackage{algorithm}
\usepackage{algorithmic}
\usepackage{subcaption}

\usepackage[misc]{ifsym}
\usepackage{adjustbox}

\usepackage{pifont}

\usepackage[dvipsnames]{xcolor}
\usepackage{color, colortbl}
\definecolor{citecolor}{HTML}{0071bc}
\definecolor{tabhighlight}{HTML}{e5e5e5}
\usepackage[colorlinks,citecolor=citecolor]{hyperref}

\makeatletter
\renewcommand\paragraph{
  \@startsection{paragraph} 
  {4} 
  {\z@} 
  {.5em \@plus1ex \@minus.2ex} 
  {-.5em} 
  {\normalfont\normalsize\bfseries} 
}
\makeatother
\begin{document}
\sloppy

\title{Rethinking Generalizability and Discriminability of Self-Supervised Learning from Evolutionary Game Theory Perspective 
}


\author{\rm Jiangmeng Li\textsuperscript{\rm 1}\!\and\!Zehua Zang\textsuperscript{\rm 1, \rm 2}\!\and\!Qirui Ji\textsuperscript{\rm 1, \rm 2}\!\and\!Chuxiong Sun\textsuperscript{\rm 1}\!\and\!Wenwen Qiang\textsuperscript{\rm 1 \textrm{\Letter}}\!\and\!Junge Zhang\textsuperscript{\rm 3}\!\and\!Changwen Zheng\textsuperscript{\rm 1}\!\and\!Fuchun Sun\textsuperscript{\rm 4}\!\and\!Hui Xiong\textsuperscript{\rm 5 \rm 6}
}




\institute{\textrm{\Letter} Wenwen Qiang \at \email{qiangwenwen@iscas.ac.cn}\\
            \textsuperscript{\rm 1} Institute of Software, Chinese Academy of Sciences, Beijing, China\\
            \textsuperscript{\rm 2} University of Chinese Academy of Sciences, Beijing, China\\
            \textsuperscript{\rm 3} Institute of Automation, Chinese Academy of Science, Beijing, China\\
            \textsuperscript{\rm 4} Tsinghua University, Beijing, China\\
            \textsuperscript{\rm 5} The Hong Kong University of Science and Technology (Guangzhou), Guangdong, China\\
            \textsuperscript{\rm 6} The Hong Kong University of Science and Technology, Hong Kong SAR, China\\
}

\date{Received: date / Accepted: date}

\maketitle


\begin{abstract}
Representations learned by self-supervised approaches are generally considered to possess sufficient \textit{generalizability} and \textit{discriminability}. However, we disclose a nontrivial \textit{mutual-exclusion} relationship between these critical representation properties through an exploratory demonstration on self-supervised learning. State-of-the-art self-supervised methods tend to enhance either generalizability or discriminability but not both simultaneously. Thus, learning representations \textit{jointly} possessing strong generalizability and discriminability presents a specific challenge for self-supervised learning. To this end, we revisit the learning paradigm of self-supervised learning from the perspective of evolutionary game theory (EGT) and outline the theoretical roadmap to achieve a desired trade-off between these representation properties. EGT performs well in analyzing the trade-off point in a \textit{two-player} game by utilizing dynamic system modeling. However, the EGT analysis requires sufficient annotated data, which contradicts the principle of self-supervised learning, i.e., the EGT analysis cannot be conducted without the annotations of the specific target domain for self-supervised learning. Thus, to enhance the methodological generalization, we propose a novel self-supervised learning method that leverages advancements in reinforcement learning to jointly benefit from the general guidance of EGT and sequentially optimize the model to chase the consistent improvement of generalizability and discriminability for specific target domains during pre-training. On top of this, we provide a benchmark to evaluate the generalizability and discriminability of learned representations comprehensively. Theoretically, we establish that the proposed method tightens the generalization error upper bound of self-supervised learning. Empirically, our method achieves state-of-the-art performance on various benchmarks. Our implementation is available at \url{https://github.com/ZangZehua/essl}.

\keywords{Self-supervised learning\and Generalizability\and Discriminability\and Evolutionary Game Theory\and Reinforcement learning}
\end{abstract}

\section{Introduction} \label{sec:intro}

As Moore's Law slows down \cite{shalf2020future}, the Connectionism-based data-driven learning paradigm hits a developmental bottleneck, since improving the model performance by simply increasing the neural network parameters is unachievable. Additionally, empowering the model's capacity by feeding large-scale annotated data to the model \cite{2009Feifei} suffers from the excessive annotating workloads. To this end, self-supervised learning emerges, developing well-designed self-supervised tasks by leveraging the guidance of specific knowledge foundations. For example, contrastive learning is based on the multi-view learning assumption \cite{2008Sridharan, 2020Tsai, wang2022chaos}, and masked language/image modeling (M-M) is based on the masking-and-reconstruction paradigm \cite{al2007learning, wang2022chaos, elkins2008six}. SSL further trains models to accomplish these self-supervised tasks in a completely unsupervised manner. Benefited from the knowledge-guided data-driven learning paradigm, SSL approaches achieve empirical successes in various fields, such as computer vision \cite{hjelm2018learning, DBLP:conf/iclr/Bao0PW22, DBLP:conf/cvpr/HeCXLDG22, DBLP:conf/cvpr/Xie00LBYD022} and natural language processing \cite{DBLP:conf/nips/BrownMRSKDNSSAA20, DBLP:conf/naacl/DevlinCLT19}.

\begin{figure}
    \centering
    \includegraphics[width=0.49\textwidth]{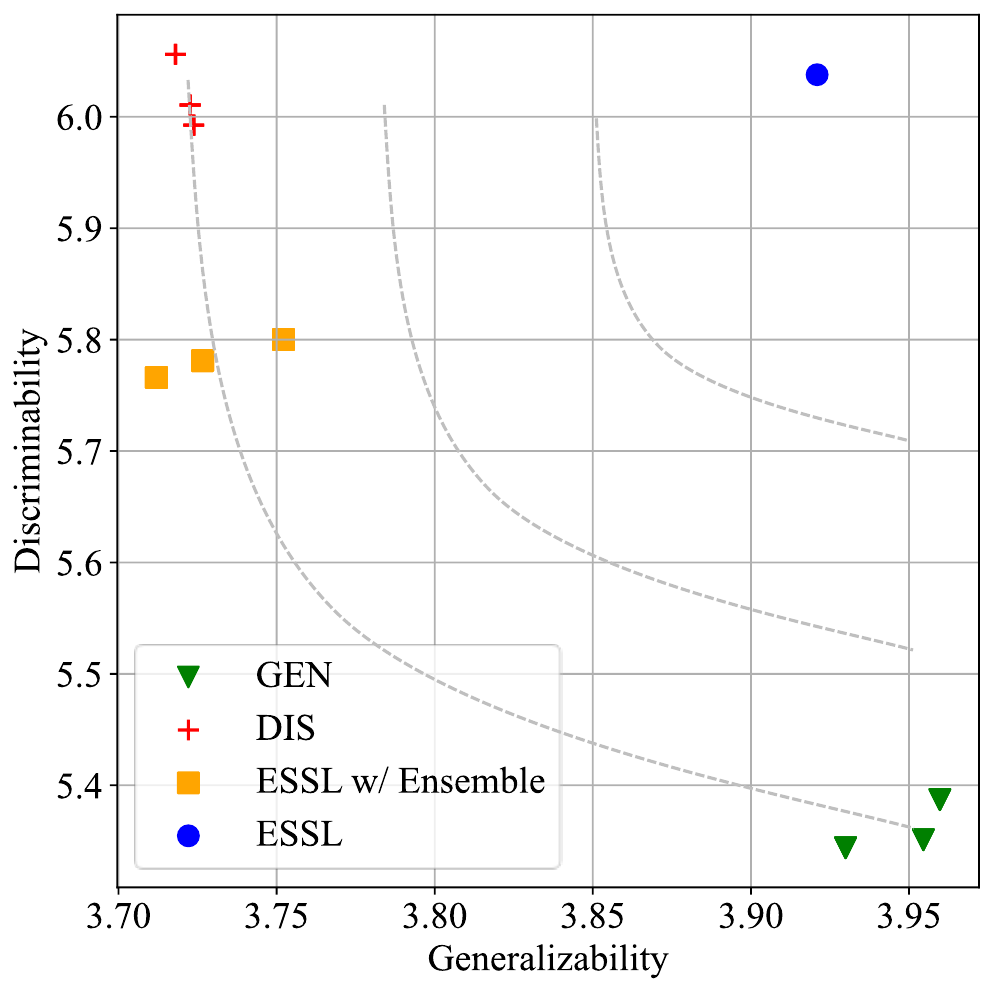}
    \caption{Generalizability and discriminability among the generalizability model (GEN), the discriminability model (DIS), ESSL w/ Ensemble, and ESSL. The generalizability model, implemented by SimCLR$^\dagger$ which is a variant of SimCLR, demonstrates better generalizability, while the discriminability model, implemented by Barlow Twins, exhibits superior discriminability. ESSL w/ Ensemble, a straightforward combination of the generalizability model and the discriminability model, results in a reduction of both generalizability and discriminability. In contrast, our proposed ESSL method effectively maintains both generalizability and discriminability simultaneously. The values of generalizability and discriminability are computed using Equation \ref{equ:gen_dis}, and the details of the four methods mentioned above can be found in \textbf{Appendix \ref{app:fig1detail}}.}
    \label{fig:motivation}
\end{figure}

In the field of SSL, the learned representations are generally regarded as having sufficient \textit{generalizability} (or \textit{transferability}) and \textit{discriminability}, and thus recent works draw limited attention to researching the intrinsic relationship between the designed self-supervised task and the properties of the representations learned by SSL methods. To this end, we investigate such representation properties for various self-supervised tasks. Specifically, we recap the preliminaries of the conventional contrastive learning paradigm: bringing views of the \textit{same} image together and pushing views of the \textit{different} images farther apart. Theoretical proofs \cite{wang2022chaos} demonstrate that such a learning paradigm encourages the model to learn the discriminative information from the inputs, but the guarantees and evidence provided by \cite{wang2022chaos, Tian2019Contrastive} further demonstrate that the contrastive objective improves the \textit{domain-specific} discriminability while degrading the \textit{cross-domain} generalizability of the learned representations. In contrast, certain SSL methods tend to boost generalizability yet undermine the discriminability of the representations, as these approaches only require representations to contain information-decoupled dimensions without the intuition to capture discriminative information \cite{2021Barlow, nakamura2000statistical, DBLP:journals/corr/abs-2210-11464}. To profoundly understand the relationship between generalizability and discriminability, we conduct a motivating experiment in Figure \ref{fig:motivation}, which demonstrates the existence of a mutual-exclusion relationship between such properties in SSL.

\textbf{Does the \textit{juste-milieu} exist to learn a self-supervised representation with strong generalizability and discriminability?}

To acquire this aspiration, we rethink the SSL paradigm from evolutionary game theory (EGT) perspective. The intuition behind our behavior is that EGT, which benefits from dynamic system modeling, can predict the rational scenarios in the game of two players or populations jointly cooperating and competing. Indirect EGT is commonly applied to examine the interaction mechanism of complex behaviors between \textit{populations} \cite{HUCK199913,JI2019116064, BESTER1998193}. On this basis, we hypothesize that fusing an SSL model with strong generalizability and another SSL model with strong discriminability in an appropriate manner can derive the desired SSL model with jointly strong generalizability and discriminability. Thus, following EGT, we treat the generalizability and discriminability models as separated populations and assign the corresponding \textit{proportions} \cite{HUCK199913,JI2019116064, BESTER1998193}, i.e., the tuning hyper-parameters. Our objective is to learn appropriate tuning hyper-parameters to boost the joint performance of generalizability and discriminability for SSL models from the perspective of EGT. Following the empirical evolutionary game analysis method \cite{gatenby2003application,BESTER1998193, HUCK199913, JI2019116064}, we define the ideal event and analyze the impacts of tuning hyper-parameters on the dynamic evolution process, thereby deriving appropriate hyper-parameters.

However, performing the complete EGT analysis on each target downstream dataset requires the over-multitudinous corresponding annotations on the target downstream dataset, which is antithetical to the principle of SSL. Thus, we propose the \textit{\underline{E}volutionary game-guided \underline{S}elf-\underline{S}upervised \underline{L}earning} approach, namely \textit{ESSL}. Specifically, we first select several annotated datasets as the \textit{representative prior} dataset, which is treated as the available inputs for the EGT analysis to generate instructive tuning hyper-parameters for the fusion of generalizability and discriminability models. Note that the selected datasets do \textit{not} include the target dataset for SSL. The intuition behind such a behavior is that the trade-off hyper-parameters derived by the EGT analysis on the representative prior datasets hold general effectiveness on various datasets. Yet, directly adopting such fixed tuning hyper-parameters cannot consistently boost the joint performance of the learned representations on specific datasets due to the distribution gaps between domains.

To this end, to learn the tuning hyper-parameters for the model fusion by jointly leveraging the guidance of EGT and dynamically adapting the hyper-parameters for the specific target dataset, the proposed ESSL introduces the reinforcement learning (RL) paradigm to enable the fitting of hyper-parameters towards the target dataset. The rationale behind the above process is that tuning the hyper-parameters in SSL is a \textit{sequential optimization} process, which adheres to the property of the Markov Decision Process, and the guidance of EGT, i.e., the instructive tuning hyper-parameters generated by the EGT analysis, can be introduced in the \textit{reward function} of RL with ease. Concretely, to comprehensively evaluate the generalizability and discriminability of learned representations, we further provide a novel self-supervised benchmark. Then, we conduct extensive comparisons on the proposed benchmark and the conventional self-supervised benchmark to sufficiently demonstrate the performance superiority of ESSL over candidate self-supervised methods. The significant contributions of this work are four-fold:

\begin{itemize}
    
    \item We disclose the mutual-exclusion relationship between the generalizability and discriminability of the representations learned by SSL methods.
    
    \item By revisiting the SSL paradigm from the perspective of EGT, we propose an innovative SSL method, namely ESSL, which jointly leverages the guidance of EGT-based analyses and the sequential optimization of the RL approach to empower representations to acquire strong generalizability and discriminability in a balanced manner.

    \item We provide theoretical analyses to demonstrate that ESSL obtains the tighter generalization error upper bound on downstream tasks compared with conventional SSL methods.

    \item Under the self-supervised benchmarks and the proposed benchmark comprehensively measuring generalizability and discriminability, the results prove the effectiveness of ESSL.
    
\end{itemize}

\section{Related Works}
{\bf{Self-supervised learning}}. In the unsupervised learning setting, self-supervised contrastive learning \cite{2020Debiased, 2020Hard, 2008Charles, 2014Marthinus, 2008Sridharan, 2020Tsai, 2020Bootstrap, 2020WhiteningErmolov, 2021Barlow, 2021Vikas, 2021Tete} has achieved great successes. Specifically, CMC \cite{Tian2019Contrastive} and AMDIM \cite{2019Philip} use contrastive learning on multi-view data. SimCLR \cite{chen2020simple} and MoCo \cite{2020Kaiming} employ large batches or memory banks to enlarge the available negative features to learn better representations. SwAV \cite{2020Mathilde} compares the cluster assignments under different views instead of directly comparing features by using more views. BYOL \cite{2020Bootstrap} and Barlow Twins \cite{2021Barlow} present a crucial issue: insufficient self-supervision may lead to feature collapse. I-JEPA \cite{DBLP:conf/cvpr/AssranDMBVRLB23} predicts the representations of various target blocks in the same image without relying on hand-crafted data augmentations, thereby improving generalizability. SiameseIM \cite{DBLP:conf/cvpr/TaoZ0HLZ00D23} shows that it is possible to achieve better discriminability by obtaining both semantic alignment and spatial sensitivity with a single dense loss. DINO \cite{DBLP:conf/iccv/CaronTMJMBJ21} discloses a significant discovery: self-supervised vision Transformers (ViTs) exhibit unique explicit information representation capabilities in image semantic segmentation. In other words, self-supervised ViTs possess stronger discriminability compared to supervised ViTs or traditional convolutional neural networks. Therefore, DINO \cite{DBLP:conf/iccv/CaronTMJMBJ21, oquab2024dinov} involves constructing a self-distillation method to train and update the teacher and student networks, employing techniques such as parameter updates similar to exponential moving averages and multiple cropping strategies for training input images. Benefiting from the well-designed architectures, benchmark SSL methods, e.g., SSL-HSIC \cite{li2021self}, IPMC \cite{DBLP:journals/nn/LiGQZ23}, MAE \cite{mae}, and CAE \cite{cae}, provide relatively sufficient self-supervision to solely improve the generalizability or discriminability performance. Orthogonal to existing methods, we focus on jointly improving the generalizability and discriminability of representations, thereby boosting the SSL model performance.

{\bf{Evolutionary game theory}}. Considering bounded rationality and learning mechanisms, EGT \cite{osborne1994course, easley2010networks, vincent1985evolutionary} focuses on the decision-making process and solves multiple equilibriums well in biology \cite{gatenby2003application} and economics \cite{BESTER1998193, HUCK199913, JI2019116064}. \cite{gatenby2003application} uses EGT to frame the tumor-host interface, elucidating key biological parameters controlling the advance of tumor tissue into the surrounding host tissue. \cite{BESTER1998193, HUCK199913} elaborate on whether altruism and fair distribution are evolutionarily stable in some instances. \cite{JI2019116064} applies EGT to examine the interaction mechanism of complex behaviors between local governments and auto manufacturers. Inspired by the practical successes of EGT, the guidance of EGT is expected to be promising in our case.

{\bf{Reinforcement learning}}. Recently, RL has achieved great successes in many real-world domains, such as Game AI \cite{silver2017mastering, silver2018general, vinyals2019grandmaster}, Robotics \cite{kalashnikov2018scalable, huttenrauch2017guided}, autonomous vehicles \cite{kiran2021deep}, and so on. Considering the strong sequential decision and policy optimization ability, RL can be regarded as a promising solution for Artificial General Intelligence (AGI). Furthermore, RL has been successfully applied in many Computer Vision (CV) and Natural Language Processing (NLP) tasks for hyper-parameter tuning \cite{fasta3rl, craftingtoolchain, DBLP:journals/tetci/KumarPS23, DBLP:journals/thms/SunYXZYDZ23}. Hence, we utilize Proximal Policy Optimization (PPO) \cite{ppo}, a well-studied policy gradient RL method, to search for the optimal hyper-parameters that balance ESSL's generalizability and discriminability. 

\label{sec:method}
\begin{figure*}[t]
    \centering
    \includegraphics[width=1.\textwidth]{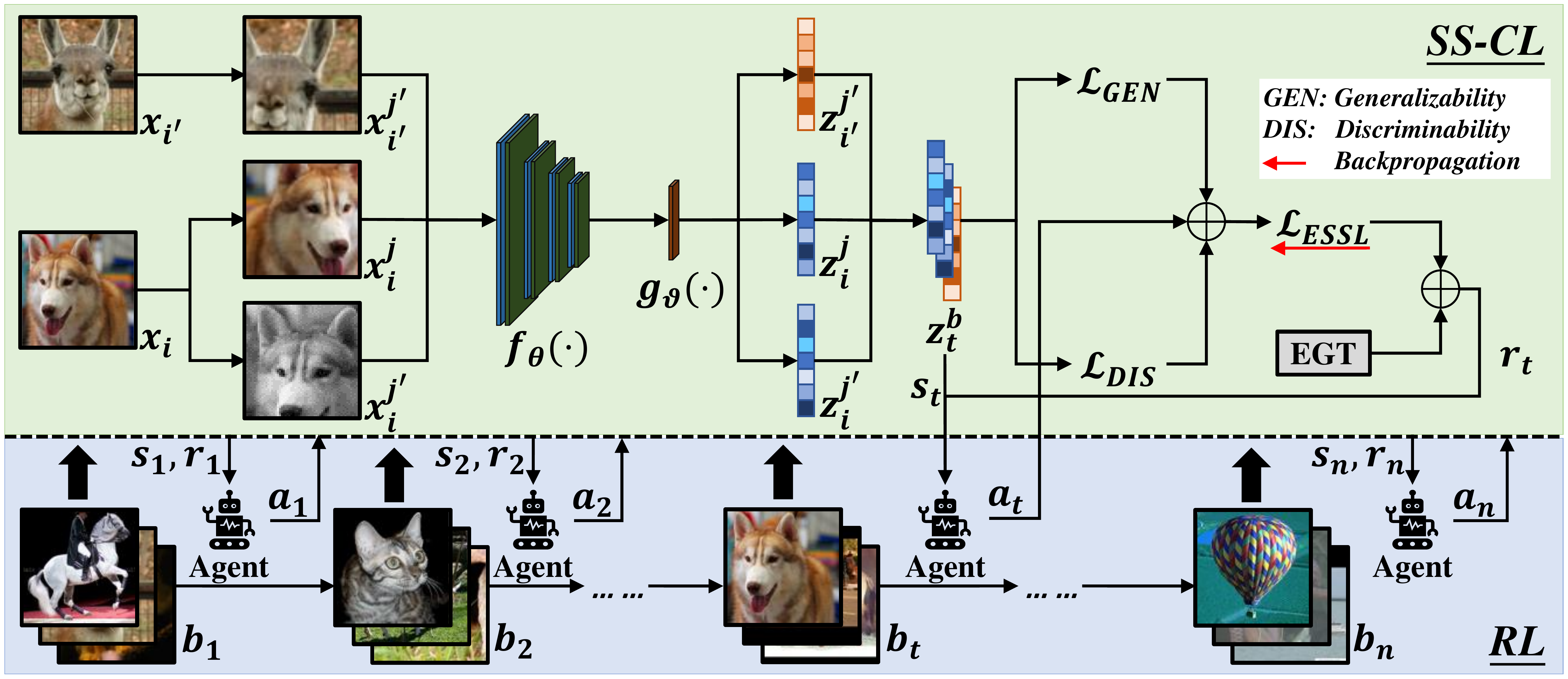}
    \vspace*{0pt}
    \caption{The framework of ESSL. The module in green denotes the training procedure of the SSL model. The module in blue is the modeled Markov Decision Processes of RL to adjust weights for the \textit{generalizability} loss, i.e., $\mathcal{L}_{GEN}$, and the \textit{discriminability} loss, i.e., $\mathcal{L}_{DIS}$, in a self-paced manner.}
    \label{fig:esslframe}
\end{figure*}

\section{Preliminaries}
\label{sec:preliminary}
To elaborate on the detailed implementation of ESSL, we recap the necessary preliminaries in this section.

\subsection{Preliminaries of Self-Supervised Learning}
{\bf{Contrastive learning}}.
Given $X = \left\{x_i^j \left| i \in \left\{1, \ldots, N^S \right\} \ and \ j \in \left\{1, \ldots, M^V\right\} \right. \right\}$, as a multi-view dataset, self-supervised contrastive learning (SS-CL) \cite{Tian2019Contrastive, chen2020simple} learns representations by maximizing agreements between the views of the same samples $\{x_i^j, x_i^{j^\prime}\}$ $j, j^\prime \in \{1,\ldots,M^{V}\}$ and $j \neq j^\prime$ (\textit{positive pairs}), while minimizing agreements between the views of different samples $\{x_i^j, x_{i^\prime}^{j^\prime}\}$ $j, j^\prime \in \{1,\ldots,M^{V}\}$ and $i \neq i^\prime$ (\textit{negative pairs}). The input data $x_i^j$ is fed into the encoder $f_{\theta}(\cdot)$ to learn a representation $h_i^j$, and $h_i^j$ is mapped into a feature $z_i^j$ by a projection head $g_{\vartheta}(\cdot)$, where $\theta$ and $\vartheta$ are the network parameters, respectively. $f_{\theta}(\cdot)$ and $g_{\vartheta}(\cdot)$ are trained by using a contrastive loss, i.e., InfoNCE \cite{2018RepresentationOord}:
\begin{equation} \label{eq:cl}
    \mathcal{L}_{InfoNCE} = -\mathbf{E}_{X_S}[\log \frac{d(\{z^+\})}{d(\{z^+\}) + \sum_{k = 1}^K d(\{z^-\}_k\})}],
\end{equation}
where $X_S$ is \textit{i.i.d.} sampled from $X$, $z^+$ denotes a positive pair, $z^-$ denotes a negative pair, and $d(\cdot)$ is a similarity measurement formula, e.g., cosine similarity.

{\bf{Feature constraint learning}}.
As another representative SSL paradigm, feature constraint learning achieves impressive performance superiority. Specifically, Barlow Twins \cite{2021Barlow} proposes a novel self-supervised feature constraint learning loss, which is formulated as follows:
\begin{equation}
    \begin{aligned}
        {\mathcal{L}_{Barlow Twins}} =& \sum_{i}(1-C_{ii})^2+\epsilon \sum_i\sum_{{i^\prime} \neq i}C_{ii^\prime}^2,\\ \ C_{ii^\prime}=&\frac{\sum_bz_{b,i}^j z_{b,i^\prime}^{j^\prime}}{\sqrt{\sum_b(z_{b,i}^j)^2}\sqrt{\sum_b(z_{b,i^\prime}^{j^\prime})^2}},
    \end{aligned}
    \label{eq:barlowtwins}
\end{equation}
where $\epsilon$ is a positive constant, trading off the importance of the first and second terms of the loss, $b$ indexes batch samples. $C_{ii^\prime}$ is the cross-correlation matrix computed between $z_{b,i}^j$ and $z_{b,i^\prime}^{j^\prime}$ with values in the range between $-1$ and $1$, and $i, i^\prime, j, j^\prime$ are indexes.  

\subsection{Preliminaries of Evolutionary Game Theory}
We recap the necessary introductions of EGT \cite{weibull1997evolutionary, easley2010networks}. The key insight of EGT is that many behaviors involve the interaction of multiple organisms in a population, and the success of any one of these organisms depends on how its behavior interacts with that of others. Thus, we evaluate the fitness of an individual organism in the context of the full population in which it lives. In EGT, fitness is like a payoff, which depends on the strategies of the organisms with which it interacts. Evolutionary Stable Strategy (ESS) and Replicator Dynamics (RD) are two critical concepts in EGT \cite{weibull1997evolutionary, easley2010networks}, which are introduced as follows.

{\bf{Evolutionary stable strategy}}.
ESS refers to a strategy in a population that cannot be replaced by any other strategy with higher fitness once it has reached a certain proportion. In other words, an ESS is a strategy that, once it comes to a certain proportion of the population, will remain stable and maintain that proportion through the evolutionary process. This concept was proposed by the biologist John Maynard Smith \cite{smith1974theory} and is now widely used in various evolutionary game models \cite{gatenby2003application, BESTER1998193, HUCK199913, JI2019116064}.

{\bf{Replicator dynamics}}.
RD is a dynamic model that describes changes in the proportion of strategies in evolutionary games. It is based on the assumption that individuals in each generation will ``replicate'' their strategy to the next generation with a certain probability, and the payoff of each strategy determines this probability. Using RD, we can predict how the proportion of strategies in evolutionary games will change over time under different initial conditions. 

The basic form of the constructed RD equation is:
\begin{equation}
    \frac{dx_i(t)}{dt} = [f(s_i,x)-f(x,x)]x_i,
\end{equation}
where $s_i$ is one strategy in strategy sets, $x_i$ denotes what proportion of individuals in the game group select the strategy $s_i$ at time $t$, $f(s_i,x)$ denotes the expected payoff of the individual $s_i$, $f(x,x)$ denotes the expected payoff for the entire population, and $dx(t)/dt$ represents the change rate of individuals that select the strategy $s_i$ per unit time. This equation can be explained as follows: the rate of change in individual strategy proportion is directly proportional to the difference between the current strategy's payoff and the average group payoff. When $f(s_i,x)-f(x,x)>0$, the proportion of strategies will increase. On the contrary, when $f(s_i,x)-f(x,x)<0$, its strategy proportion will decrease.


\section{Methodology}
In this section, we elaborate on the proposed ESSL, and the approach framework is illustrated in Figure \ref{fig:esslframe}. The detailed nomenclature is provided in \textbf{Appendix \ref{app:nomenclature}}. We emphasize the roadmap of ESSL as follows: 1) we transform the target objective of achieving a SSL representation with strong generalizability and discriminability into the target objective of tuning the hyper-parameters of two decoupled SSL methods and further combining the candidate methods to build the desired SSL method in an ensemble manner; 2) we perform the EGT analysis for the generalizability and discriminability of SSL methods on \textit{certain} representative prior datasets to derive the average hyper-parameters, which is treated as the guidance for the following RL-based hyper-parameter tuning process; 3) we impose the RL-based approach to jointly leverage the guidance of the EGT analysis and tackle the sequential optimization of the hyper-parameter tuning. Therefore, the derived ESSL can be well generalized into various SSL fields.

\begin{table*}
    \renewcommand\arraystretch{1.3}
    \caption{Payoff matrix for both players, i.e., the generalizability and discriminability models, and the payoffs of players are \textit{different}. Refer to \textbf{Section \ref{sec:egt ans}} for the denotations.}
    \label{tab:payoffsep}
    \setlength{\tabcolsep}{25.pt}
     \begin{center}
        \begin{tabular}{cccc}
            \toprule
            & & \multicolumn{2}{c}{Discriminability model}\\
            \cline{3-4}
            & & $A$ & $U$ \\
            \hline
            \multirow{3}*{Generalizability model} & \multirow{2}*{$A$} & $\omega_1$ $\cdot$ $($ $G_1$ - $N_2$ $)$ + $($ $D_2$ - $N_1$ $)$; & $\omega_1$ $\cdot$ $G_1$ + $D_1$; \\
            & & $($ $G_1$ - $N_2$ $)$ + $\omega_2$ $\cdot$ $($ $D_2$ - $N_1$ $)$ & $G_1$ + $\omega_2$ $\cdot$ $D_1$ \\
            \cline{2-4}
            & $U$ & $\omega_1$ $\cdot$ $G_2$ + $D_2$; $G_2$ + $\omega_2$ $\cdot$ $D_2$ & $0$; $0$ \\
            \bottomrule
        \end{tabular}
    \end{center}
\end{table*}

\subsection{Revisiting the Learning Paradigm of the Evolutionary Game Theory Perspective}
\label{sec:egt ans}
As the motivating exploration in Figure \ref{fig:motivation}, we demonstrate the negative correlation between the generalizability and discriminability of a representation learned by SSL. Concretely, to chase the trade-off between the generalizability and discriminability, we intuitively revisit the learning paradigm of self-supervised approaches from the EGT perspective and thus treat the generalizability and discriminability models as two \textit{populations} working against and cooperating. Precisely, the process of tuning the hyper-parameters, i.e., the impacts of the generalizability and discriminability models on the fusion of such models \textit{exactly fits} the case of population evolution in EGT, and the dynamic property of the hyper-parameter tuning process adheres to the requirement of EGT, such that we state that adopting EGT to derive the desired trade-off between generalizability and discriminability is theoretically sound.

{\bf{EGT model assumption}}.
The current benchmark cannot comprehensively evaluate the generalizability and discriminability of a self-supervised model, which vary due to different pretext tasks or loss functions, such that following the procedure of EGT, we propose an \textit{integrated benchmark}, and such a benchmark further is the foundational \textit{assumption} of EGT model. Specifically, assuming that supervised learning (SL) achieves lower classification errors than SSL on the downstream tasks, we propose a benchmark to measure the generalizability and discriminability of self-supervised models. Given two datasets, $\mathcal{D}$ and $\mathcal{D}^\prime$, and a self-supervised model $M$, based on the pre-training on $\mathcal{D}$, $G$ denoted the generalizability of $M$ on $\mathcal{D}^\prime$, and $D$ denotes the discriminability of $M$ on $\mathcal{D}$:
\begin{equation}\label{equ:gen_dis}
    \begin{aligned}[c]
        G =& \frac{1}{ACC_{SL}(\mathcal{D}^\prime)-ACC_{M}(\mathcal{D} \rightarrow \mathcal{D}^\prime)},\\ D =& \frac{1}{ACC_{SL}(\mathcal{D})-ACC_{M}(\mathcal{D} \rightarrow \mathcal{D})},
    \end{aligned}
\end{equation}
where $ACC_{SL}(\mathcal{D})$ and $ACC_{SL}(\mathcal{D}^\prime)$ denote the classification accuracies on $\mathcal{D}$ and $\mathcal{D}^\prime$, respectively. $ACC_{M}(\mathcal{D} \rightarrow \mathcal{D})$ and $ACC_{M}(\mathcal{D} \rightarrow \mathcal{D}^\prime)$ denote the accuracy on $\mathcal{D}$ and $\mathcal{D}^\prime$ pre-trained on $\mathcal{D}$. In this way, higher values of D or G indicate better discriminability or generalizability of the model.

Given a generalizability model $M_g$ and a discriminability model $M_d$, $M_{ens}$ denotes the ensemble model of $M_g$ and $M_d$. We denote $N_1$ as the negative impact of $M_g$ on the discriminability of $M_d$, and $N_2$ denotes the negative impact of $M_d$ on the generalizability of $M_g$:
\begin{equation}
    \begin{aligned}
        N_1 &= ACC_{SL}(\mathcal{D}^\prime)-ACC_{M_{ens}}(\mathcal{D} \rightarrow \mathcal{D}^\prime),\\ N_2 &= ACC_{SSL}(\mathcal{D} \rightarrow \mathcal{D}^\prime)-ACC_{M_{ens}}(\mathcal{D} \rightarrow \mathcal{D}^\prime),
    \end{aligned}
\end{equation}
where $ACC_{M_{ens}}(\mathcal{D} \rightarrow \mathcal{D}^\prime)$ denotes the accuracy on $\mathcal{D}^\prime$ pre-trained by the ensemble model on $\mathcal{D}$.

\begin{figure}
    \centering
    \includegraphics[width=0.49\textwidth]{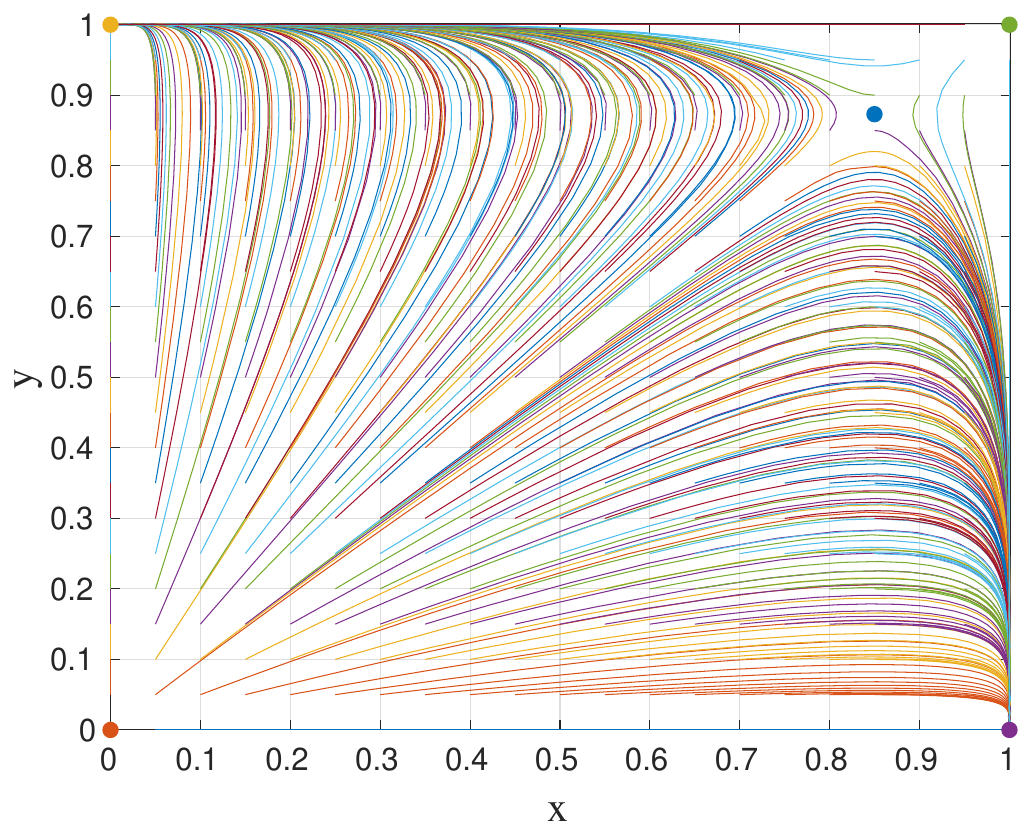}
    \caption{Phase diagram for the dynamic system. $x$ and $y$ represent the proportion of discriminability and generalizability models, respectively. Every line denotes the evolving strategy of the candidate models under different initialization.}
    \label{fig:phase diagram}
\end{figure}

\begin{figure}
    \centering
    \includegraphics[width=0.49\textwidth]{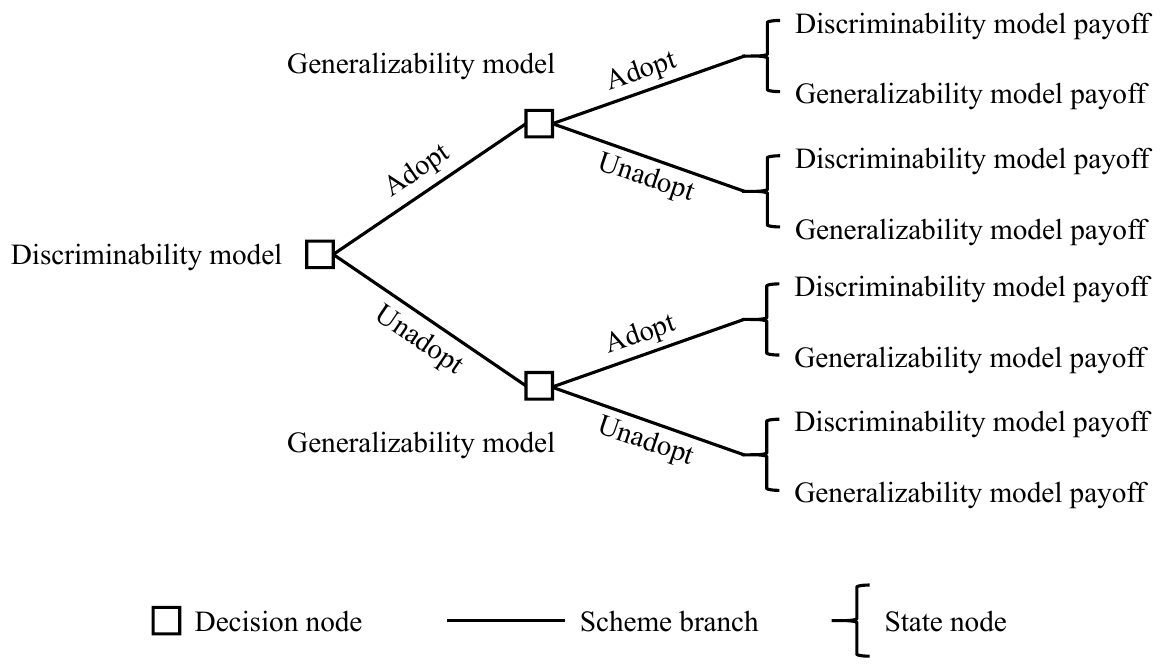}
    \caption{Decision tree illustration for the EGT model between generalizability and discriminability.}
    \label{fig:decisiontree}
\end{figure}
\textbf{EGT model framework}. We develop a \textit{group-double population indirect evolutionary model} \cite{HUCK199913,JI2019116064, BESTER1998193}, where the generalizability and discriminability models present the sufficient and finite populations, respectively, of the chosen pure strategies. The strategy spaces of both players are $S_G = \{ A, U \}$ and $S_D = \{ A, U \}$, where $A$ denotes ``Adopted'' and $U$ denotes ``Unadopted''. The indirect EGT can be expressed as follows: the combined model selects models based on individual utility maximization. Note that the utility contains both generalizability and discriminability. The system state changes along with the time, and the model utility changes accordingly, resulting in sequential and dynamical adjustments in the strategies chosen by generalizability and discriminability models. Accordingly, the decision tree for the EGT model between generalizability and discriminability is provided in Figure \ref{fig:decisiontree} for ease of understanding.

Based on the above assumption, Table \ref{tab:payoffsep} illustrates a payoff matrix for players across different strategies. In this table, $G_1$ and $D_1$ represent the generalizability and discriminability metrics for $M_g$, and $G_2$ and $D_2$ correspond to $M_d$. $\omega_1$ and $\omega_2$ indicate the weights assigned to generalizability and discriminability.

\textbf{EGT model analysis}.
Let $x$ $\left(0 \leq x \leq 1\right)$ represent the proportion of discriminability models, and $y$ $(0 \leq y \leq 1)$ represents the proportion of generalizability models which select the $A$ strategy. On the contrary, $1-x$ and $1-y$ represent the proportion of discriminability and generalizability models selecting the $U$ strategy, respectively. $\bf{H}$ denotes the income matrix of the generalizability model, which is defined by
\begin{equation}
    \bf{H} = \begin{pmatrix}
        \omega_1 \cdot (G_1 - N_2)+(D_2-N_1) & \omega_1 \cdot G_1+D_1 \\
        \omega_1 \cdot G_2 + D_2 & 0
    \end{pmatrix}.
\end{equation}
When the generalizability model selects the strategy $A$, given $\bf{e}$, $\bf{x}$ and $\bf{y}$ denoting the proportion metrics of the chosen strategy, respectively, the expected utility is defined by
\begin{equation}
    \begin{aligned}
        &U_{G1}={\bf{eHx^{\top}}}=\begin{pmatrix}
        1 & 0
    \end{pmatrix} \cdot \\ &\begin{pmatrix}
        \omega_1 \cdot (G_1 - N_2)+(D_2-N_1) & \omega_1 \cdot G_1+D_1 \\
        \omega_1 \cdot G_2 + D_2 & 0
    \end{pmatrix} \begin{pmatrix}
        x \\ 1-x
    \end{pmatrix},
    \end{aligned}
\end{equation}
and the corresponding average expected utility of the generalizability models is implemented by
\begin{equation}
    \begin{aligned}
        &\overline{U_G} = {\bf{yHx^{\top}}} = \begin{pmatrix}
        y & 1-y
    \end{pmatrix} \cdot \\ &\begin{pmatrix}
        \omega_1 \cdot (G_1 - N_2)+(D_2-N_1) & \omega_1 \cdot G_1+D_1 \\
        \omega_1 \cdot G_2 + D_2 & 0
    \end{pmatrix} \begin{pmatrix}
        x \\ 1-x
    \end{pmatrix}.
    \end{aligned}
\end{equation}
According to the Malthusian equation \cite{malthus1872essay}, the growth of $A$ action of the generalizability model is equivalent to $U_{G1}-\overline{U_G}$. 
Specifically, when $U_{G1} > \overline{U_G}$, it means that the utility of taking action $A$ is higher than the average utility of the group, so accordingly, the proportion of this strategy will grow. Vice versa, i.e., when $U_{G1} < \overline{U_G}$ or $U_{G1} = \overline{U_G}$, the proportion of this strategy will decrease or remain unchanged.
Therefore, the RD equation is defined as follows:
\begin{equation}
\label{eq:fy}
    \begin{aligned}
        &F(y) = \frac{dy}{dt}=y[{\bf{eHx^{\top}}}-{\bf{yHx^{\top}}}] = y(1-y) \cdot \\
        &[\omega_1\cdot G_1+D_1-(D_1+\omega_1\cdot 	G_2+\omega_1\cdot N_2+N_1)x],
    \end{aligned}
\end{equation}
which denotes the change rate of generalizability models that select the strategy $A$ per unit time.
In detail, the factor $y(1-y)$ represents the current proportion change rate as related to the proportion ratios of the strategies choosing $A$ and $U$, 
and the term $[\omega_1\cdot G_1+D_1-(\cdot \cdot \cdot)x]$ represents the mutual influence and competitive effects of strategy $A$ against other strategies.
The \textit{first derivative} of Equation 9 represents the \textit{acceleration or deceleration} of the \textit{growth rate} of strategy $A$ in generalizability models, formally expressed as:
\begin{equation}
\label{eq:fy'}
    \begin{aligned}
        &\frac{dF(y)}{dt} = (1-2y) \ \cdot \\ &[\omega_1\cdot G_1+D_1-(D_1+\omega_1\cdot G_2+\omega_1\cdot N_2+N_1)x].
    \end{aligned}
\end{equation}
It can be easily demonstrated that $y=0$ or $y=1$, and $x=x^\star$ are the numerical solutions of the equation $F(y)=dy/dt=0$, where 
\begin{equation} \label{eq:xstar}
    x^\star=\dfrac{\omega_1\cdot G_1+D_1}{D_1+\omega_1 \cdot G_2+\omega_1\cdot N_2+N_1}.
\end{equation}
Under the firm condition that $0\leq x^\star \leq 1$, based on the stability theorem \cite{friedman1991evolutionary}, we can derive that when $F(x)=0$ and $F^\prime(x)\leq 0$, the proposition that $x$ is the ESS holds. For any $y$, if $x=x^\star$, $F(y)=0$, and $F^\prime(y)=0$ are satisfied, axis $y$ is in a stable state. With the conditions that $x<x^\star$, $F^\prime(y)|_{y=0}>0$, and $F^\prime(y)|_{y=1}<0$, we indicate that $y=1$ is the only ESS. Accordingly, if $x>x^\star$, $F^\prime(y)|_{y=0}<0$, and $F^\prime(y)|_{y=1}>0$ holds, we derive that $y=0$ is the only ESS.

Adhering to the aforementioned principle, we can obtain the income matrix $\bf{K}$ of the discriminability model and the corresponding RD equation, which are defined as follows:
\begin{equation}
    \bf{K} = \begin{pmatrix}
        (G_1 - N_2)+\omega_2 \cdot (D_2-N_1) & G_1+\omega_2 \cdot D_1 \\
        G_2 + \omega_2 \cdot D_2 & 0
    \end{pmatrix},
\end{equation}
and
\begin{equation}
\label{eq:fx}
    \begin{aligned}
        &F(x)=\dfrac{dx}{dt}=x[{\bf{eK^{\top}y^{\top}-xK^{\top}y^{\top}}}]=x(1-x) \ \cdot \\ &[G_2+\omega_2\cdot D_2-(\omega_2 \cdot D_1+G_2+N_2+\omega_2\cdot N_1)y].
    \end{aligned}
\end{equation}
Equation \ref{eq:fx} represents the change rate of discriminability models that select the strategy $A$.
The \textit{first derivative} of Equation \ref{eq:fx} means the \textit{change rate} of the \textit{growth rate} of strategy $A$ in discriminability models, detailed as follows:
\begin{equation}
\label{eq:fx'}
    \begin{aligned}
        &\frac{dF(x)}{dt} 
         = (1-2x) \cdot \\ &[G_2+\omega_2\cdot D_2-(\omega_2\cdot 						D_1+G_2+N_2+\omega_2\cdot N_1)y],
    \end{aligned}
\end{equation}
where $x=0$ or $x=1$, and $y=y^\star$ are the numerical solutions of the equation $F(x)=dx/dt=0$, and
\begin{equation} \label{eq:ystar}
    y^\star=\dfrac{G_2+\omega_2\cdot D_2}{\omega_2\cdot D_1+G_2+\omega_2 \cdot N_1+N_2}.
\end{equation}

With the condition that $0\leq y^\star \leq 1$, we obtain that for any x, if $y=y^\star$, $F(x)=0$, and $F^\prime(x)=0$ holds, axis $x$ is in a stable state. If $y<y^\star$, $F^\prime(x)|_{x=0}>0$, and $F^\prime(x)|_{x=1}<0$ are satisfied, $x=1$ is the only ESS. Accordingly, if $y>y^\star$, $F^\prime(x)|_{x=0}<0$, and $F^\prime(x)|_{x=1}>0$ are satisfied, $x=0$ is the only ESS.

According to Equation \ref{eq:fy} and Equation \ref{eq:fx}, we obtain a 2-dim nonlinear dynamic system for the generalizability and the discriminability models, which formulates as follows:
\begin{equation}
\label{eq:dynamic system}
\left\{  
    \begin{aligned}
        F(x)=\dfrac{dx}{dt} 
                = & x(1-x)[G_2+\omega_2\cdot D_2\\ & -(\omega_2\cdot 						D_1+G_2+N_2+\omega_2\cdot N_1)y] \\  
        F(y)=\dfrac{dy}{dt} 
                = & y(1-y)[\omega_1\cdot G_1+D_1\\ & -(D_1+\omega_1\cdot 					G_2+\omega_1\cdot N_2+N_1)x]     
    \end{aligned}
\right.,
\end{equation}

\begin{table}[t]
    \setlength{\tabcolsep}{10pt}
    \renewcommand\arraystretch{1.08}
    \centering
    \caption{Local stability analyses of equilibrium points.}
    \label{tab:local stability analyses}
    \begin{tabular}{cccc}
    \hline 
    point & $det(J)$ & $tr(J)$ & Result \\
    \hline
    (0,0) & + & + & unstable point  \\
    (0,1) & + & - & stable point \\
    (1,0) & + & - & stable point \\
    (1,1) & + & + & unstable point \\
    ($x^\star$,$y^\star$) & - & 0 & saddle point \\
    \hline
    \end{tabular}
    \label{tab:my_label}
\end{table}

{\bf{Local stability analyses of equilibrium points}}.
\label{app:proof the local stability analyses}
According to Equation \ref{eq:dynamic system}, when $dx/dt=0$ or $dy/dt=0$, the system reaches an equilibrium. Concretely, the equilibrium points of the dynamic system are $(0,0)$, $(0,1)$, $(1,0)$, $(1,1)$, $(x^\star,y^\star)$. 
According to Friedman's proposal, i.e., ``describe (co)evolution of population(s) with dynamics defined by differential equations'' \cite{friedman1991evolutionary}, we obtain the Jacobian matrix $J$ of Equation \ref{eq:dynamic system} as follows:
\begin{equation}
    \begin{aligned}
        J=&\begin{bmatrix}
            \partial F(x)/\partial x & \partial F(x)/\partial y \\
            \partial F(y)/\partial x & \partial F(y)/\partial y 
        \end{bmatrix} \\
        =&\begin{bmatrix}
            (1-2x) \cdot [G_2+\omega_2\cdot D_2 & 
            -x(1-x) \cdot \\
            -(\omega_2\cdot D_1 +G_2 & 
            (\omega_2\cdot D_1+G_2 \\
            +N_2+\omega_2\cdot N_1)y] &
            +N_2+\omega_2\cdot N_1) \\ \\
            -y(1-y) \cdot &
            (1-2y) \cdot [\omega_1\cdot G_1+D_1 \\
            (D_1+\omega_1\cdot G_2 &
            -(D_1+\omega_1\cdot G_2 \\
            +\omega_1\cdot N_2+N_1) &
            +\omega_1\cdot N_2+N_1)x]
        \end{bmatrix}.
    \end{aligned}
\end{equation}
From Table \ref{tab:local stability analyses}, we can get results that the equilibrium points $(0,0),(1,1)$ are unstable points; $(0,1),(1,0)$ are stable points; $(x^\star,y^\star)$ is a saddle point.
Figure \ref{fig:phase diagram} demonstrates the dynamic system's phase diagram. Intuitively, we reckon that $(x^\star,y^\star)$ can be a compromised point for the combined model to obtain the satisfactory trade-off between the generalizability and discriminability. Such a compromise point $(x^\star,y^\star)$ can be derived by adopting the approach, i.e., Equation \ref{eq:xstar} and Equation \ref{eq:ystar}. The experiments in \textbf{Section \ref{sec:expegt}} empirically prove the inference behind our behavior.


\subsection{Chasing the Trade-off between Generalizability and Discriminability}
In practice, we notice that conducting the EGT-based analyses on each downstream task incurs excessive workloads, which require significant human annotation and computational costs. Additionally, solely generalizing the hyper-parameters derived from the EGT analysis on several representative downstream tasks cannot achieve the desired model performance improvement. The corresponding empirical proofs are provided in \textbf{Section \ref{sec:experiment}}. Therefore, we first perform the EGT analysis on the representative prior datasets to derive the average hyper-parameters, serving as the guidance of EGT. Then, we propose to leverage a practical RL approach to adjust the hyper-parameters dynamically within the guidance of EGT. Refer to \textbf{Section  \ref{sec:experiment}} for the setting of representative prior datasets in practice.

{\bf{Performing the reinforcement learning}}.
Given a multi-view dataset $X$, we propose to learn a self-supervised representation with the desired trade-off between the generalizability and discriminability. In practice, we intuitively leverage an \textit{ensemble} manner, which can be formally described as follows:
\begin{equation} \label{eq:essl}
    \mathcal{L}_{ESSL}=\alpha \mathcal{L}_{GEN} + \beta \mathcal{L}_{DIS},
\end{equation}
where $\alpha$ and $\beta$ are the hyper-parameters controlling the balance between $\mathcal{L}_{DIS}$ and $\mathcal{L}_{GEN}$, i.e., the losses present the generalizability and discriminability, respectively. In the implementation, we adopt $\mathcal{L}_{InfoNCE}$ as $\mathcal{L}_{GEN}$ and $\mathcal{L}_{Barlow Twins}$ as $\mathcal{L}_{DIS}$. To chase the best trade-off, we employ an effective RL method, PPO \cite{ppo}. At each time step $t$, the RL agent receives a state $s_t$, which contains information the agent can observe. In our formulation, the state is formulated by
\begin{equation}
    s_t=F_s(z_t^b),
\end{equation}
where $F_s(\cdot)$ is the mean function, and $z_t^b$ is the features of batch $b$. Under state $s_t$, the agent takes an action $a_t = (a_t^\alpha, a_t^\beta)$. Then, the parameters $\alpha$ and $\beta$ at the time step $t$ can be calculated as:
\begin{equation}
    \alpha_t=V+a_t^\alpha, \ \beta_t=V+a_t^\beta,
\end{equation}
where $V$ is the hyper-parameter.

{\bf{Employing the guidance of EGT}}.
Given a set of representative datasets $S_d=\{\mathcal{D}^1, \mathcal{D}^2,\ldots, \mathcal{D}^n\}$, we primarily derive the average result of the SL accuracies on such datasets, and refer to \textbf{Appendix \ref{app:slr}} for the detailed results. Accordingly, $\overline{G}$ denotes the average SSL accuracy measuring the generalizability, $\overline{D}$ denotes the average SSL accuracy measuring the discriminability, $\overline{N_1}$ denotes the average effect on the discriminability model by the generalizability model, and $\overline{N_2}$ denotes the average effect on the generalizability model by the discriminability model.

In our framework, the \textit{reward function} is composed of two key components:
\begin{equation} \label{eq:rewardf}
    r_t= G(\alpha_t, \beta_t, x^\star, y^\star) + \xi\frac{1}{|\mathcal{L}_{{ESSL}_t}-\phi \mathcal{L}_{{ESSL}_{t-1}}|},
\end{equation}
The first term is defined by $ G(\cdot,\cdot,\cdot,\cdot)$, representing the cosine similarity function. Here, $(x^\star, y^\star)$ denotes the \textit{a prior} balanced point between generalizability and discriminability, as determined by EGT. The hyper-parameter $\xi$ is utilized to balance different reward signals, while $\phi$ is another hyper-parameter designed for the exploration term. Overall, the first term of the reward function allows RL to dynamically pursue a balance between generalizability and discriminability based on prior knowledge acquired from general datasets. However, relying solely on this general prior knowledge often falls short in uncovering the latent characteristics specific to a dataset. Hence, we have devised a second term in the reward function, aimed at encouraging agents to explore the most suitable balance point for the specific dataset.


\section{Theoretical Insights on the Proposed ESSL}\label{sec:theory}
In this section, we introduce various theoretical analyses, encompassing: 1) from the perspective of generalization theory, we demonstrate that the effectiveness and generalization of the proposed ESSL are consistently superior to the state-of-the-art SSL methods; 2) from the causality perspective, we prove that our proposed learning mechanism, determining the suitable balancing hyper-parameters for combining the generalizability and discriminability models in a self-paced manner, is reasonable, which empowers the candidate SSL model to dynamically chase the optimal trade-off between generalizability and discriminability.

\subsection{Theoretical Analyses upon Generalization Error Bounds}
Benchmarking models on classification tasks is a standard evaluation approach for self-supervised approaches. Therefore, we present the generalization error bounds of the proposed ESSL on classification tasks. The benchmarking is implemented by training a softmax classifier to minimize the traditional cross-entropy loss \cite{zhang2018generalized} as follows:
\begin{equation} \label{eq:softmax}
    {\mathcal{L}_{softmax}}( {f;T} ) = {\inf _W} \ {\mathcal{L}_{cross-entropy}}( {W \cdot f;T} ),
\end{equation}
where $f$ denotes the fixed encoder pre-trained by self-supervised methods, $W$ is the linear probing classifier, and $T$ is the label on the downstream task. Holding Equation \ref{eq:softmax}, we present the generalization error for the feature representation $f(X)$ learned from $X$ by
\begin{equation} \label{eq:gesm}
     \mathcal{L}_{GE}^{f} = {\mathbf{E}_X}\left[ {\inf _W} \ {\mathcal{L}_{cross-entropy}}\left( {W \cdot f\left(X\right);T} \right)\right].
\end{equation}
For the encoder $f$, we can further bridge the generalization error $\mathcal{L}_{GE}^{f}$ with the conventional contrastive learning objective $\mathcal{L}_{InfoNCE}^f$, defined in Equation \ref{eq:cl}, by introducing the \textit{Rademacher complexity} \cite{2019Arora, DBLP:conf/icml/QiangLZ0X22} as follows:

\begin{theorem}
\label{thm:bigtheorem}
Suppose ${f^\star}$ is a sufficiently trained neural network by following the proposed ESSL objective and following Equation \ref{eq:essl}, ${f^\star}$ is acquired as follows:
\begin{equation} \label{eq:gesm1}
     f^\star = \mathop{\arg\min}_f \alpha \mathcal{L}_{GEN}^f + \beta \mathcal{L}_{DIS}^f.
\end{equation}
Given the condition that $\alpha$ is frozen, with the probability at least $1 - \delta$, we hold that the upper bound of generalization error can be approximated by
\begin{equation} \label{eq:ge}
    \begin{aligned}
        \left| {\mathcal{L}_{GE}^{f^\star} - \mathcal{L}_{InfoNCE}^{f^\star}} \right| \le & O\Bigg( \frac{{{\sqrt {1 + {1 \mathord{\left/  {\vphantom {1 N^{batch}}} \right. \kern-\nulldelimiterspace} N^{batch}}}}\cdot{\mathcal{R}_{\mathcal{H}}}\left( \beta  \right)}}{{N^{S}}} \\ & + \sqrt {\frac{{{\log \left( {{1 \mathord{\left/  {\vphantom {1 \delta }} \right. \kern-\nulldelimiterspace} \delta }} \right) \cdot {\log ^2}\left( {N^{S}} \right)}}}{{N^{S}}}} \Bigg),
    \end{aligned}
\end{equation}
where ${N^{S}}$ denotes the number of training samples, and $N^{batch}$ denotes the size of a mini-batch. ${{\mathcal{R}_{\mathcal{H}}}\left( \beta  \right)}$ presents the Rademacher complexity, which establishes a constant property that ${{\mathcal{R}_{\mathcal{H}}}\left( \beta  \right)}$ is monotonically decreasing w.r.t. $\beta$. 
\end{theorem}

We provide proof to attest to the correctness and integrity of the proposed theoretical analyses. To demonstrate the validity of the corresponding Theorem, we provide a Lemma as follows:

\begin{lemma}
\label{f:1}
Let $f \in \mathcal{H}$, where $\mathcal{H}$ is a restricted function hypothesis space, and ${f^\star}$ be a neural network sufficiently trained by following Equation \ref{eq:gesm}. When $\alpha$ is fixed, with the probability at least $1 - \delta$ over the training data $X = \left\{ {{X_1},{X_2},\ldots,{X_{N^{S}}}} \right\}$, the following inequality holds:
\begin{equation} \label{eq:geproof1}
    \begin{aligned}
        \mathcal{L}_{GE}^{f^\star} \le \mathcal{L}_{InfoNCE}^{f^\star} + &O\Bigg(\frac{{{\sqrt {1 + {1 \mathord{\left/  {\vphantom {1 N^{neg}}} \right. \kern-\nulldelimiterspace} {N^{S}}}}} \cdot{\mathcal{R}_{\mathcal{H}}}\left( \beta  \right)}}{{N^{S}}} \\ &+ \sqrt {\frac{{{\log \left( {{1 \mathord{\left/  {\vphantom {1 \delta }} \right. \kern-\nulldelimiterspace} \delta }} \right) \cdot {\log ^2}\left( {N^{S}} \right)}}}{{N^{S}}}} \Bigg).
    \end{aligned}
\end{equation}
The Rademacher Complexity is defined as
\begin{equation} \label{eq:geproofrc}
    \mathcal{R}_{\mathcal{H}}\left( \beta  \right) = \mathop{\mathbf{E}} \limits_{\sigma  \in {{\left\{ { \pm 1} \right\}}^{3\cdot d \cdot N^S}}} \left[ {{\mathop{\sup}\limits_{f \in \mathcal{H}_\beta}}\left\langle {\sigma ,f} \right\rangle } \right],
\end{equation}
where $d$ denotes the dimensionality of the representations learned by $f$, and the restricted function hypothesis set $\mathcal{H}_\beta$ corresponding to the coefficient $\beta$ of the proposed ESSL is determined by
\begin{equation}
    \mathcal{H}_\beta = \left\{ {f \left| {f  \in \mathcal{H},{\ \rm{and }} \ {{\cal R}_1}\left( f \right) \le {4 \mathord{\left/
    {\vphantom {4 \beta }} \right.
    \kern-\nulldelimiterspace} \beta }} \right.} \right\}.
\end{equation}

\end{lemma}

Holding Lemma \ref{f:1}, we can thus derive

\begin{corollary}
\label{cor:1}
Following the conventional contrastive principle, for ${f^\star}$, with the probability at least $1 - \delta$, we have that
\begin{equation} \label{eq:geproof2}
    \begin{aligned}
        \left| {\mathcal{L}_{GE}^{f^\star} - \mathcal{L}_{InfoNCE}^{f^\star}} \right| \le &O\Bigg(\frac{{{\sqrt {1 + {1 \mathord{\left/  {\vphantom {1 N^{neg}}} \right. \kern-\nulldelimiterspace} N^{neg}}} }\cdot{\mathcal{R}_{\mathcal{H}}}\left( \beta  \right)}}{{N^{S}}} \\ &+\sqrt {\frac{{{\log \left( {{1 \mathord{\left/  {\vphantom {1 \delta }} \right. \kern-\nulldelimiterspace} \delta }} \right) \cdot {\log ^2}\left( {N^{S}} \right)}}}{{N^{S}}}} \Bigg),
    \end{aligned}
\end{equation}
where ${N^{S}}$ is the total number of training samples, and $N^{neg}$ is the size of negative pairs, which is greatly close to ${N^{S}}$ for the integrated dataset due to the instance-level contrastive principle. ${{\mathcal{R}_{\mathcal{H}}}\left( \beta  \right)}$ is the Rademacher complexity, and ${{\mathcal{R}_{\mathcal{H}}}\left( \beta  \right)}$ is monotonically decreasing w.r.t. $\beta$.
\begin{proof}
    Holding Equation \ref{eq:softmax} and Equation \ref{eq:gesm}, for the generalization error $\mathcal{L}_{GE}^{f}$, we have:
    \begin{equation}
    \begin{aligned}
        &\mathcal{L}_{GE}^{f}\\
        = \ &{\mathbf{E}_X}\left[ {\mathop {\inf }\limits_W {\mathcal{L}_{cross-entropy}}\left( {W \cdot f,T} \right)} \right]\\        
        = \ &{\mathbf{E}_X}\left[ { - \log \frac{{{e^{f{{\left( X \right)}^T}{u_{{c^{pos} }}}}}}}{{{e^{f{{\left( X \right)}^T}{u_{{c^{pos} }}}}} + \sum {{e^{f{{\left( X \right)}^T}{u_{{c^{neg} }}}}}} }}} \right]\\
        = \ &{\mathbf{E}_X}\Bigg[ - \log \bigg({{{e^{f{{\left( X \right)}^T}{u_{{c^{pos} }}}}}}}\bigg) \\ & \ \ \ \ \ \ + \log \bigg({{{e^{f{{\left( X \right)}^T}{u_{{c^{pos} }}}}} + \sum {{e^{f{{\left( X \right)}^T}{u_{{c^{neg} }}}}}} }}\bigg) \Bigg]\\
        = \ &{\mathbf{E}_X}\Bigg[ \log \bigg({e^{f{{\left( X \right)}^T}{\mathbf{E}_{{X^{pos} }}}\left[ {f\left( {{X^{pos} }} \right)} \right]}} \\ & \ \ \ \ \ \ + {N^{S}}\mathbf{E}\left[ {{e^{f{{\left( X \right)}^T}{E_{{X^{neg} }}}\left[ {f\left( {{X^{neg} }} \right)} \right]}}} \right]\bigg) \\ & \ \ \ \ \ \ - \log\bigg( {{e^{f{{\left( X \right)}^T}{\mathbf{E}_{{X^{pos} }}}\left[ {f\left( {{X^{pos} }} \right)} \right]}}}\bigg) \Bigg]\\
        \ge \ &{\mathbf{E}_X}\Bigg[ \log \bigg({e^{f{{\left( X \right)}^T}{\mathbf{E}_{{X^{pos} }}}\left[ {f\left( {{X^{pos} }} \right)} \right]}} \\ & \ \ \ \ \ \ + {N^{S}}{\mathbf{E}_{{X^{neg} }}}\left[ {{e^{f{{\left( X \right)}^T}{\mathbf{E}_{{X^{neg} }}}\left[ {f\left( {{X^{neg} }} \right)} \right]}}} \right]\bigg) \\ & \ \ \ \ \ \ - \log\bigg( {{e^{f{{\left( X \right)}^T}{\mathbf{E}_{{X^{pos} }}}\left[ {f\left( {{X^{pos} }} \right)} \right]}}}\bigg) \Bigg]\\
        = \ &\mathcal{L}_{InfoNCE}^{f}.
    \end{aligned}
    \end{equation}
    Then, we can intuitively obtain
    \begin{equation}
    \begin{aligned}
        &\mathcal{L}_{GE}^{f^\star} - \mathcal{L}_{InfoNCE}^{f} \\
        \le \ &\mathcal{L}_{InfoNCE}^{f^\star} - \mathcal{L}_{InfoNCE}^{f} \\
        \le \ &O\Bigg(\frac{{{\sqrt {1 + {1 \mathord{\left/  {\vphantom {1 N^{neg}}} \right. \kern-\nulldelimiterspace} N^{neg}}} }\cdot{\mathcal{R}_{\mathcal{H}}}\left( \beta  \right)}}{{N^{S}}} \\ &+ \sqrt {\frac{{{\log \left( {{1 \mathord{\left/  {\vphantom {1 \delta }} \right. \kern-\nulldelimiterspace} \delta }} \right) \cdot {\log ^2}\left( {N^{S}} \right)}}}{{N^{S}}}} \Bigg).
    \end{aligned}
    \end{equation}
    Therefore, Corollary \ref{cor:1} holds.
\end{proof}
\end{corollary}
Specifically, we apply Corollary \ref{cor:1} in practical scenarios, where only a mini-batch of training samples is available for the neural network encoder $f$ in a gradient back-propagating step, such that a tractable surrogate variable is introduced in Equation \ref{eq:geproof2}, i.e., $N^{neg}$ is converted into $N^{batch}$. Therefore, we theoretically substantiate the correctness and integrity of Theorem \ref{thm:bigtheorem}.

Considering Equation \ref{eq:ge}, we observe that the upper bound of generalization error gradually decreases with the increase of the training sample size ${N^{S}}$. Such an observation is consistent with the conventional SL methods. On top of this, for the relatively non-trivial mini-batch size $N^{batch}$, the first term of generalization error ${{\sqrt {1 + {1 \mathord{\left/  {\vphantom {1 N^{batch}}} \right. \kern-\nulldelimiterspace} N^{batch}}}} \cdot{\mathcal{R}_{\mathcal{H}}}\left( \beta  \right)}$ is effectively reduced. Thus the generalization error upper bound is tightened, consistent with the empirical induction in SS-CL approaches. Connecting the proposed approach with the generalization error, we attest that the enlargement of the coefficient $\beta$ of ESSL directly results in the reduction of the Rademacher complexity $\mathcal{R}_{\mathcal{H}}$. Concretely, due to the non-triviality of $\beta$, we can theoretically substantiate that performing SSL in the ESSL manner derives a relatively tighter generalization error upper bound, such that ESSL holds stronger generalizability than the conventional methods.

\begin{figure}
	\centering
        \includegraphics[width=0.35\textwidth]{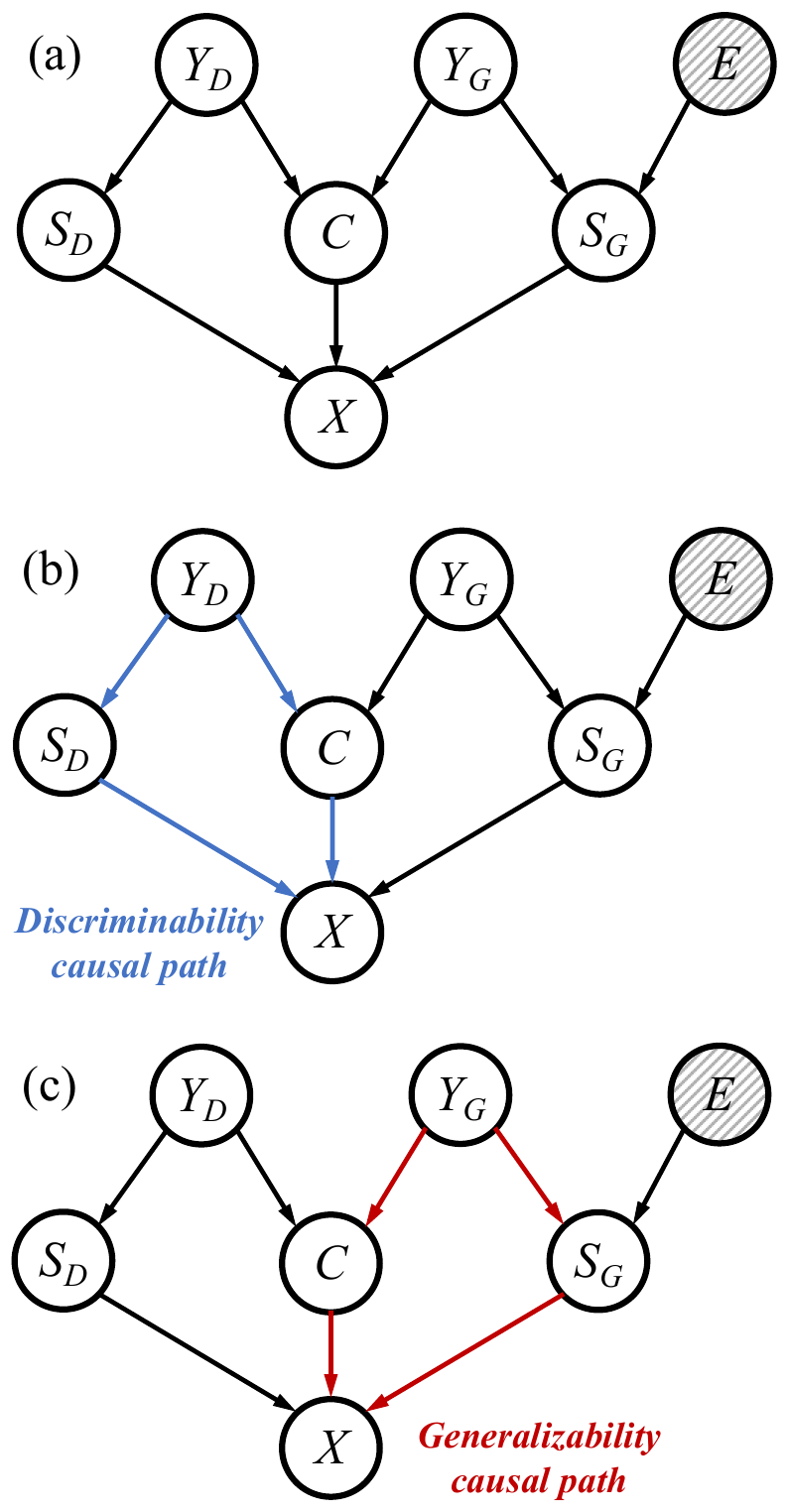}
	\caption{The proposed SCM for modeling the causality encompassing the generalizability and discriminability of SSL from the perspective of data generalization, which is illustrated in (a). (b) and (c) are the SCMs highlighting the discriminability causal path and the generalizability causal path, respectively.}
	\label{fig:scm}
	\vspace{-0.2cm}
\end{figure}

\subsection{Theoretical Analyses upon Structural Causal Model}
\textbf{Structural causal model constructions}.
We intuitively conceptualize the vanilla \textit{data generation process} of SSL with respect to generalizability and discriminability, which is achieved by introducing the structural causal model (SCM) \cite{pearl2009causality, pearl2016causal}. Inspired by \cite{2021Contrastive, 2017Nonlinear}, we determine that in a realistic context, data can be conceptualized as being derived from underlying generating factors. The process of feature extraction can thus be viewed as the extraction of these corresponding generating factors. In essence, if the feature representation of a sample can be obtained through an invertible transformation of its generating factor, then the feature representation and the generating factor of the sample can be deemed equivalent. Formally, let $Y_D$ denote the ground-truth label of the domain-specific scenario, which is strictly associated with the discriminability, and $Y_G$ presents the ground-truth label of cross-domain scenarios concerning the generalizability. $C$ denotes the invariant \textit{causal} feature, which is domain-agnostic. $S_D$ and $S_G$ denote the \textit{spurious} feature corresponding to the domain-specific scenario and cross-domain scenarios, respectively. The intrinsic reason behind the appearance of $S_D$ and $S_G$ is that besides the domain-shared feature, i.e., the invariant causal feature, different domains generally possess their \textit{exclusive} features (or patterns). The discriminability corresponds to a specific domain, while the generalizability involves multiple domains, such that $S_G$ is associated with an exogenous (unobservable) variable $E$ \cite{pearl2016causal}, depicting the environments of different domains for generalization. $X$ represents the available samples for SSL. Concretely, the data generalization process of SSL can be abstracted as follows: given the discriminability-related label $Y_D$ and generalizability-related label $Y_G$, we obtain the domain-shared invariant causal feature $C$, and the spurious features $S_D$ and $S_G$, corresponding to different domains. Note that, $S_G$ is further influenced by the environmental variable $E$. Based on the derived features, the sample data of SSL can be generated. According to the depicted process, we propose the SCM illustrated in Figure \ref{fig:scm}.


\textbf{Clarifying the mutual-exclusion between generalizability and discriminability via SCM}.
According to the proposed SCM, as demonstrated in Figure \ref{fig:scm} (a), the objective of achieving the optimal trade-off between the generalizability and discriminability of SSL models can be transformed into \textit{jointly deriving $P\left(Y_D|X\right)$ and $P\left(Y_G|X\right)$, i.e., acquiring the causation between $X$, $Y_D$, and $Y_G$}. Adhering to the causal theory, the confounding effects do \textit{not} exist in the SCM, such that we can directly derive $P\left(Y_D|X\right)$ and $P\left(Y_G|X\right)$ via fitting a nonlinear projection based on the available data. The discriminability causal path is shown in Figure \ref{fig:scm} (b), containing two ways: $Y_D \to S_D \to X$ and $Y_D \to C \to X$. The generalizability causal path is shown in Figure \ref{fig:scm} (c), also containing two ways: $Y_G \to S_G \to X$ and $Y_G \to C \to X$. We observe that $C$ is shared between causal paths, which is domain-agnostic, while $S_D$ and $S_G$ are inconsistent and domain-dependent. The fitting of the nonlinear projection is based on the available \textit{domain-dependent} data. Accordingly, the information of $S_D$ and $S_G$, captured by the representations learned via the nonlinear projection, is relatively sufficient, and thus the information entropy of $C$ is inherently limited. Concretely, for the discriminability causal path, the domain is determined, and the training data and test data are i.i.d. sampled from the same distribution, so that the information entropy of $S_D$ is limited, since the domain-specific spurious feature $S_D$ \textit{only} corresponds to a determined domain. Learning the causation from the discriminability causal path is practically available. However, for the generalizability causal path, the domains are variant, and the domain-specific spurious feature $S_G$ corresponds to multiple domains, which is influenced by the environmental variable $E$, and thus the information entropy of $S_G$ is significantly large, exacerbating the estimation of causation from the causal path. An intuitive explanation is that in cross-domain scenarios of generalizability, the training data and test data are \textit{not} i.i.d. sampled from a shared distribution, and the training data is sampled from a determined domain, while the test data is sampled from different domains. In general, $P\left(Y_D|X\right)$ and $P\left(Y_G|X\right)$ cannot be accurately estimated simultaneously. In other words, generalizability and discriminability establish a certain mutual-exclusion relationship from the SCM-based theoretical perspective.

\begin{table}
    \setlength{\tabcolsep}{6.5pt}
    \renewcommand\arraystretch{1.3}
    \caption{Top-1 accuracies (in \%) with linear probing on STL10, dubbed by S10, CIFAR10, dubbed by C10, CIFAR100, dubbed by C100, and Tiny-ImageNet, dubbed by Tiny. The compared methods are pre-trained and then evaluated on conventional benchmark datasets in an iterative manner. Note that BT denotes Barlow Twins.}
    \centering
    \adjustbox{max width=\linewidth}{
    \begin{tabular}{llllll}
        \toprule
        \multirow{2}*{\textbf{PT}}          & \multirow{2}*{Model}     & \multicolumn{4}{c}{\textbf{Eval}}                                 \\\cline{3-6}
                                            &                          & S10           & C10           & C100          & Tiny              \\
        \hline
        \multirow{9}*{\rotatebox{90}{S10}}  & SimCLR                   & 76.4          & 56.0          & 28.0          & 24.6              \\   
                                            & SimCLR$^\dagger$         & 84.1          & 75.2          & 44.7          & 33.1              \\
                                            & MoCo-v2                  & 80.1          & 67.3          & 37.0          & 27.8              \\
                                            & Barlow Twins             & 85.5          & 70.0          & 41.5          & 29.8              \\
                                            & BYOL                     & 83.7          & 58.7          & 32.2          & 29.2              \\
                                            & SwAV                     & 85.1          & 74.4          & 44.0          & 32.4              \\
                                            & DINO                     & 85.8          & 75.2          & 44.8          & 33.1              \\
                                            \cline{2-6}
                                            & ESSL(SimCLR$^\dagger$+Barlow Twins)& 86.1          & 73.4          & 45.4          & 33.7    \\  
                                            & ESSL(SwAV+Barlow Twins)            & 87.8          & 76.8          & 45.7          & 35.3    \\
                                            & ESSL(DINO+Barlow Twins)            & \bf88.4       & \bf77.8       & \bf46.4       & \bf35.7 \\
        \hline
        \multirow{9}*{\rotatebox{90}{C10}}  & SimCLR                   & 62.2          & 73.4          & 42.1          & 23.8              \\   
                                            & SimCLR$^\dagger$         & 64.8          & 74.4          & 45.8          & 26.1              \\
                                            & MoCo-v2                  & 65.3          & 65.7          & 37.2          & 21.8              \\
                                            & Barlow Twins             & 73.1          & 83.0          & 43.8          & 23.1              \\
                                            & BYOL                     & 74.8          & 50.6          & 24.0          & 21.3              \\
                                            & SwAV                     & 73.6          & 79.6          & 45.3          & 26.2              \\
                                            & DINO                     & 73.6          & 80.3          & 45.4          & 26.9              \\
                                            \cline{2-6}
                                            & ESSL(SimCLR$^\dagger$+Barlow Twins)& 73.8          & \bf84.3       & 44.1          & 22.8    \\ 
                                            & ESSL(SwAV+Barlow Twins)            & 73.8          & 80.6          & 46.9          & \bf27.6 \\
                                            & ESSL(DINO+Barlow Twins)            & \bf75.3       & 81.9          & \bf48.2       & \bf27.6 \\
        \hline
        \multirow{9}*{\rotatebox{90}{C100}} & SimCLR                   & 62.2          & 73.4          & 42.0          & 25.6              \\  
                                            & SimCLR$^\dagger$         & 69.2          & 74.8          & 48.5          & 25.5              \\ 
                                            & MoCo-v2                  & 62.0          & 60.4          & 43.2          & 19.7              \\
                                            & Barlow Twins             & 69.5          & 73.8          & 51.1          & 23.7              \\
                                            & BYOL                     & 66.5          & 50.7          & 26.2          & 23.5              \\
                                            & SwAV                     & 70.0          & 75.4          & 49.4          & 25.2              \\
                                            & DINO                     & 70.2          & 75.5          & 50.2          & 25.9              \\
                                            \cline{2-6}
                                            & ESSL(SimCLR$^\dagger$+Barlow Twins)& 70.0          & 75.1          & 51.1          & 23.2    \\       
                                            & ESSL(SwAV+Barlow Twins)            & 70.5          & \bf77.2       & \bf51.8       & 25.2    \\
                                            & ESSL(DINO+Barlow Twins)            & \bf70.8       & 76.2          & \bf51.8       & \bf26.9 \\
        \hline
        \multirow{9}*{\rotatebox{90}{Tiny}} & SimCLR                   & 69.8          & 60.4          & 35.6          & 31.1              \\  
                                            & SimCLR$^\dagger$         & 75.2          & 70.6          & 40.5          & 32.6              \\ 
                                            & MoCo-v2                  & 67.3          & 57.6          & 27.6          & 21.3              \\
                                            & Barlow Twins             & 76.0          & 70.0          & 39.9          & 32.7              \\
                                            & BYOL                     & 73.0          & 58.0          & 31.6          & 32.3              \\
                                            & SwAV                     & 75.5          & 71.3          & 40.9          & 32.9              \\
                                            & DINO                     & 76.2          & 71.9          & 41.7          & 33.5              \\
                                            \cline{2-6} 
                                            & ESSL(SimCLR$^\dagger$+Barlow Twins)& 76.8          & 73.4          & 40.5          & 32.0    \\       
                                            & ESSL(SwAV+Barlow Twins)            & 76.7          & \bf74.2       & \bf43.2       & 35.5    \\
                                            & ESSL(DINO+Barlow Twins)            & \bf78.1       & 73.8          & 41.9          & \bf36.6 \\
        \bottomrule
    \end{tabular}
    }
    \label{tab:main_exp_4}
\end{table}

\textbf{A theoretical explanation of the proposed trade-off mechanisms}.
Inspired by the information bottleneck theory \cite{DBLP:conf/iclr/Federici0FKA20, alemi2016deep, tishby2000information} and data processing inequality \cite{1991Elements}, we determine that from the perspective of Markov process, the nonlinear projection-based data processing is a process of a gradual decrease in information entropy. Accordingly, compared with the \textit{whole} input image space, the information entropy of the learned representation space is widely lower, and thus, the representation \textit{cannot} comprehensively contain all information of $C$, $S_D$, and $S_G$. To chase the expected trade-off between the generalizability and discriminability, we propose to sufficiently model the information of $C$ in the learned representation instead of $S_D$ and $S_G$. However, for the existing SSL tasks, acquiring $C$ is entangled with the processes of acquiring $S_D$ or $S_G$, e.g., solely enhancing the generalizability of SSL model jointly acquires $C$ and $S_G$, and solely enhancing the discriminability of SSL model jointly acquires $C$ and $S_D$. Therefore, balancing the acquisition of $C$ and $S_G$ and the acquisition of $C$ and $S_D$ to maximize the modeling of $C$ is the essence of chasing the trade-off between the generalizability and discriminability. On top of this, due to the variance of domains for generalization, the information entropy of $S_G$ and $S_D$ is \textit{unbalanced}, i.e., the information entropy of $S_G$ is significantly larger than $S_D$, and the information entropy of $S_D$ and $S_G$ is inconsistent among different training domains. During balancing the enhancement of generalizability and discriminability for SSL models, plainly adopting the equivalent or firm hyper-parameters as the weights is theoretically and practically suboptimal. In contrast, by dynamically leveraging the EGT analysis and RL-based hyper-parameter optimization approach, our proposed ESSL can sufficiently model $C$ during SSL training. Concretely, from the joint viewpoint of causality and information theory, we substantiate the theoretical soundness of the intrinsic mechanism behind the proposed ESSL.

\section{Experiments} \label{sec:experiment}
In this section, we comprehensively conduct a series of experiments to evaluate the proposed ESSL from multiple perspectives. In this regard, we bootstrap the evaluation by answering the following research questions:
\begin{enumerate}
    \item Is ESSL consistently competitive with other baselines on conventional benchmark datasets (Section: \ref{sec:expsmall}) and the large-scale benchmark dataset (\textbf{Section \ref{sec:explarge}})?
    \item How about the generalizability and discriminability of self-supervised methods with different loss functions on various datasets (\textbf{Section \ref{sec:expgen_dis}})?
    \item How can we combine the two loss functions as best as possible from the network structure perspective (\textbf{Section: \ref{sec:expprojnum}})?
    \item How much performance improvement does the imposed EGT bring (\textbf{Section \ref{sec:expegt}})?
    \item How does RL shift the balance achieved by the EGT analysis (\textbf{Section \ref{sec:expsaddle}})?
\end{enumerate}

\textbf{Experimental schemes}. In practice, the generalizability is measured by conducting the pre-training and testing on \textit{heterogeneous} datasets, which is determined as the generalization task. In contrast, the discriminability is measured by conducting the pre-training and testing on the \textit{homogeneous} dataset, which is treated as the discrimination task.

Note that the representative prior datasets include STL10 \cite{stl10}, CIFAR10 \cite{tinyimagenetcifar10100}, CIFAR100 \cite{tinyimagenetcifar10100}, and Tiny-ImageNet \cite{tinyimagenetcifar10100} for the detailed implementation of the EGT analysis.

\textbf{Benchmark baselines}. The compared baselines include SimCLR \cite{chen2020simple}, MoCo \cite{2020Kaiming}, CMC \cite{Tian2019Contrastive}, CPC \cite{henaff2020data}, InfoMin Aug \cite{tian2020makes}, MoCo-v2 \cite{2020Kaiming}, SwAV \cite{2020Mathilde}, Barlow Twins \cite{2021Barlow}, BYOL \cite{2020Bootstrap}, SSL-HSIC \cite{li2021self}, MAE \cite{mae}, VICReg \cite{vicreg}, SwAV \cite{swav}, DINO \cite{oquab2024dinov}, CAE \cite{cae}, MV-MR \cite{mvmr}, and GroCo \cite{groco}. Note that, for conventional benchmark experiments, the compared methods consistently leverage ResNet-18 \cite{resnet} as the backbone networks; for large-scale benchmark experiments, most candidate methods, including the proposed ESSL, adopt standard ResNet-50 \cite{resnet} as the backbone network, while MAE and CAE adopt ViT-B-32 \cite{vit} as the backbone network. In detail, ResNet-50 possesses 25M parameters, and ViT-B possesses 86M parameters. On top of this, MAE and CAE impose 1600 epochs during training, while the rest of the methods impose 1000 epochs.

\subsection{Performance of Classification Task on Conventional Benchmark Datasets}\label{sec:expsmall}
ESSL is a combination of the generalizability model and discriminability model to jointly achieve better generalizability and discriminability. Based on our experimental results, the generalizability model has alternative options, encompassing SimCLR$^\dagger$, SwAV, and DINO. The discriminability model is consistently implemented by Barlow Twins.

Through an image classification task, we evaluate the generalizability and discriminability of ESSL on four conventional benchmark datasets, i.e., STL10 \cite{stl10}, CIFAR10 \cite{tinyimagenetcifar10100}, CIFAR100 \cite{tinyimagenetcifar10100}, and Tiny-ImageNet \cite{tinyimagenetcifar10100}. Note that computing generalizability and discriminability requires the detailed implementation of baselines, such that we compare the proposed ESSL with the most representative benchmark methods in the field of SSL for the joint evaluation of generalizability and discriminability. Based on our experimental results, the generalizability model has alternative options SimCLR$^\dagger$, SwAV and DINO, while the discriminability model is implemented by Barlow Twins. SimCLR$^\dagger$ is a variant of SimCLR, adapted to match the settings of Barlow Twins. This adjustment is made to ensure the fairness in evaluating the impact of different loss functions on generalizability and discriminability, which is the primary focus of our method. By aligning the settings, we can isolate the effect of the loss function, thereby preventing other factors from influencing the final experimental results. We pre-train ResNet-18 as the backbone and three linear layers as the projection head for 500 epochs on each dataset for each method as the pre-training phase. Then, we use the fixed backbone pre-trained above and further train a linear layer on every dataset as the linear probing phase. More training details are in \textbf{Appendix \ref{app:hyper-param}}. The top-1 classification accuracy of each method on each dataset is reported in Table \ref{tab:main_exp_4}. All the experimental results are achieved using the code in \textbf{Appendix \ref{app:coderepo}}, and we collect the results of 10 trials and derive the average results for comparisons.

\begin{figure}
    \centering
    \vspace*{0pt}
    \includegraphics[width=0.49\textwidth]{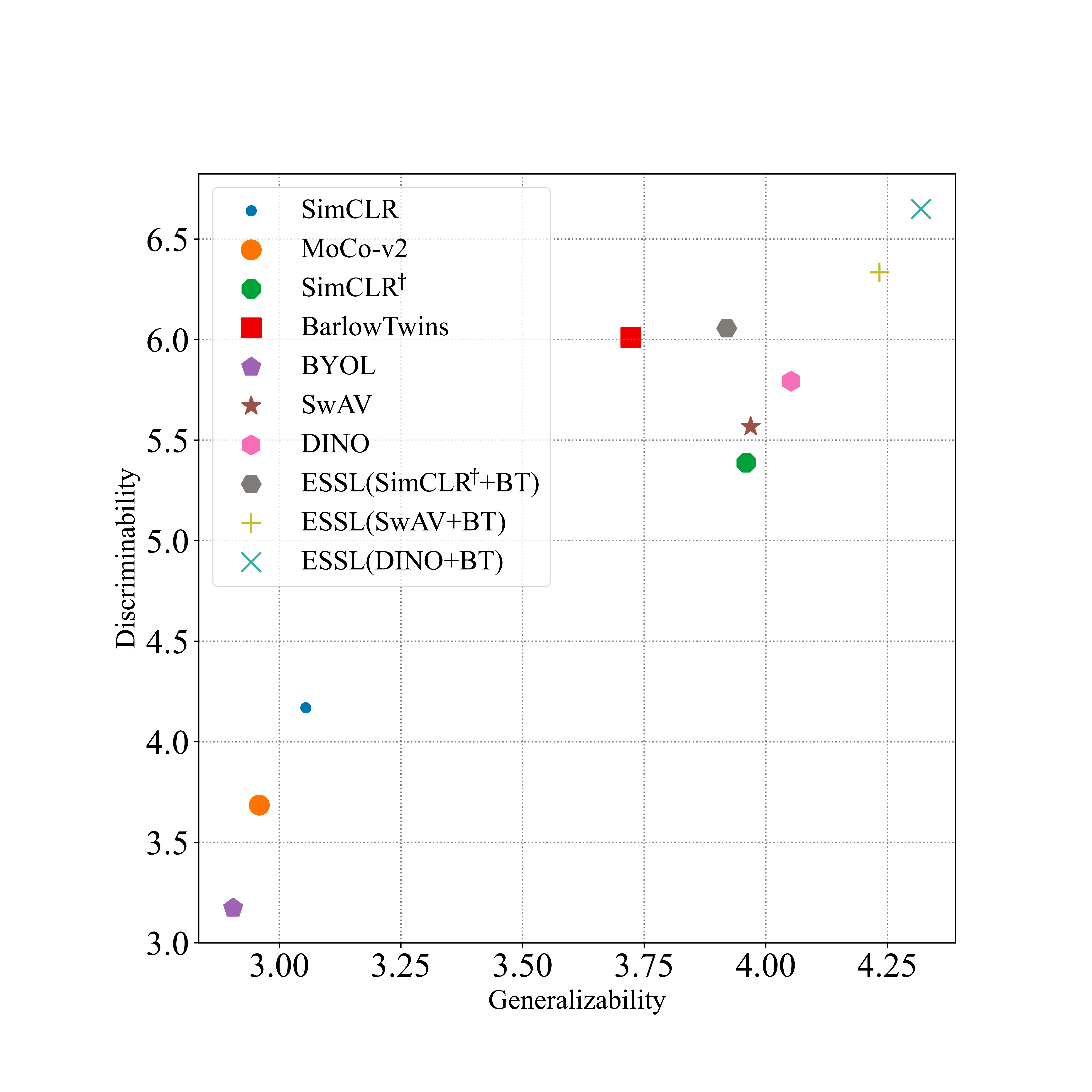}
    \caption{Evaluating generalizability and discriminability on the proposed EGT-based benchmark.}
    \label{fig:gd}
\end{figure}

Overall, our proposed ESSL achieves state-of-the-art performance with respect to both generalizability and discriminability. Specifically, when implementing DINO as the generalizability model and Barlow Twins as the discriminability model, ESSL(DINO+Barlow Twins) achieves the best performance across all datasets. Additionally, the other two implementations, ESSL(SimCLR$^\dagger$+Barlow Twins) and ESSL(SwAV+Barlow Twins), also demonstrate higher generalizability and discriminability compared to the baselines. This indicates that ESSL can effectively learn representations that excel in both generalizability and discriminability simultaneously. To better visualize these attributes, we follow the metric proposed in \textbf{Section \ref{sec:egt ans}}, with the results presented in Figure \ref{fig:gd}. The numerical values of the experimental results are multiplied by 100 to facilitate observation and avoid ambiguous decimals. These findings align with the motivating experiments, confirming that ESSL consistently achieves the best overall performance.

\subsection{Performance of Classification Task on Large-Scale Benchmark Dataset}\label{sec:explarge}
\textbf{Discriminability evaluation on ImageNet}. To further demonstrate the performance of our method, we evaluate ESSL(SimCLR$^\dagger$+BT) and ESSL(DINO+BT) on the large-scale benchmark dataset, ImageNet \cite{2009Feifei}, using 1000 epochs of pre-training followed by 100 epochs of linear probing. From the discriminability perspective, the experimental results are presented in Table \ref{tab:main_exp_imagenet}. Additional implementation details are provided in \textbf{Appendix \ref{app:impdetails}}. Note that all baseline results are taken directly from the corresponding published papers. As shown in Table \ref{tab:main_exp_imagenet}, our proposed ESSL outperforms all baselines in the image classification task, indicating that ESSL can effectively learn representations with strong discriminability on a large-scale benchmark dataset. Furthermore, exchanging SimCLR$^\dagger$ to DINO, which has higher generalizability, can further enhance ESSL.


\begin{figure*}
    \centering
    \begin{subfigure}[t]{0.3\textwidth}
        \centering
        \includegraphics[width=1.0\linewidth]{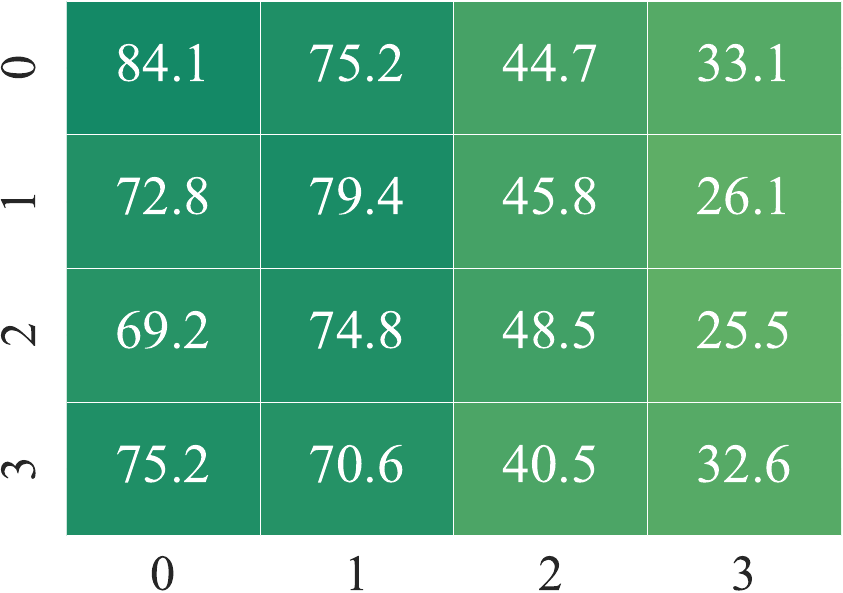}
        \caption{SimCLR$^\dagger$.}
    \end{subfigure}%
    \quad
    \begin{subfigure}[t]{0.3\textwidth}
        \centering
        \includegraphics[width=1.0\linewidth]{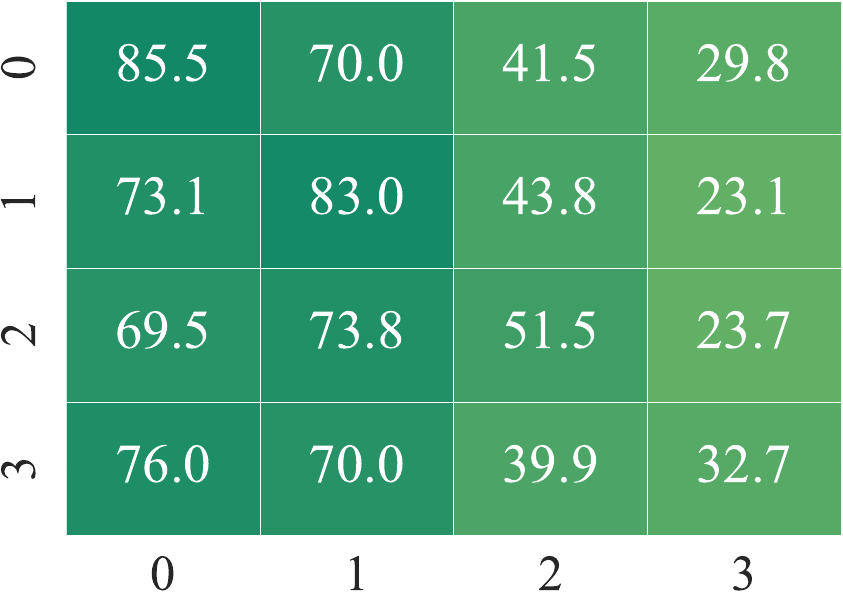}
        \caption{Barlow Twins.}
    \end{subfigure}%
    \quad
    \begin{subfigure}[t]{0.29\textwidth}
        \centering
        \includegraphics[width=1.0\linewidth]{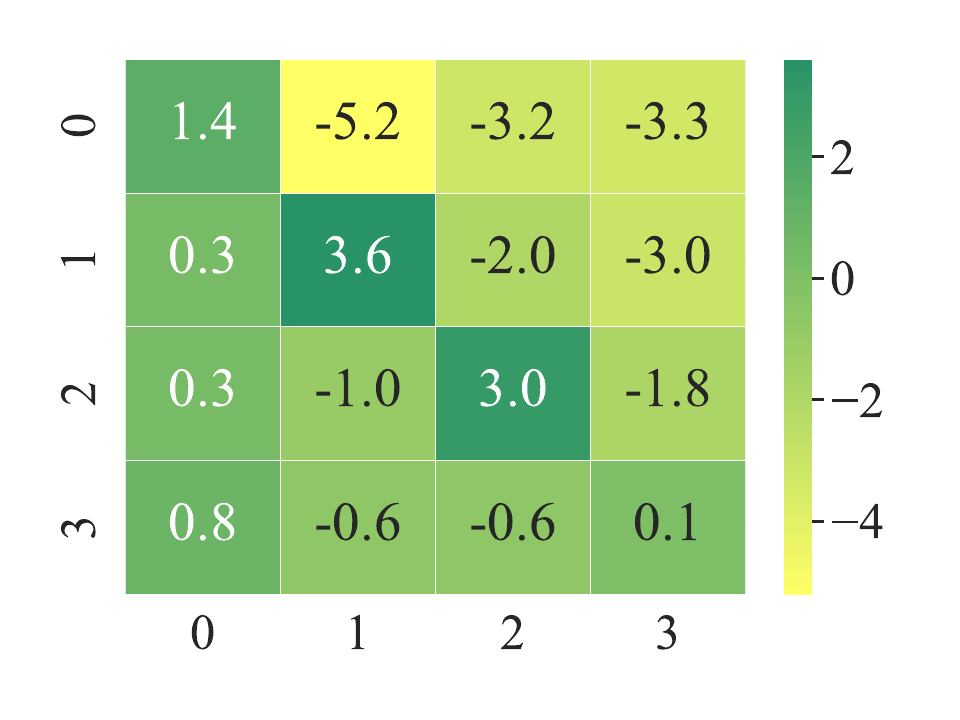}
        \caption{Residual.}
    \end{subfigure}
    \caption{Generalizability and discriminability evaluations with different self-supervised losses. (a) shows the results pre-trained with $\mathcal{L}_{GEN}$ for stronger generalizability. (b) shows the results pre-trained $\mathcal{L}_{DIS}$ for stronger discriminability. (c) demonstrates the residuals by subtracting (a) from (b). All the results are under the same training paradigm.}
    \label{fig:inbt}
\end{figure*}

\textbf{Generalizability evaluation with imposing the transfer learning tasks on ImageNet}. We evaluate the learned representations of ESSL on additional classification datasets to assess whether the representations learned on ImageNet can be generalized to heterogeneous domains, thus demonstrating the generalizability of ESSL. Using the pre-trained models of ESSL(SimCLR$^\dagger$+Barlow Twins) and ESSL(DINO+Barlow Twins) from \textbf{Section \ref{sec:explarge}}, we perform linear evaluation on the same set of classification tasks used in \cite{chen2020simple, 74}, carefully adhering to their evaluation protocols. The performance results are reported in Table \ref{tab:main_exp_transfer}, where ESSL(SimCLR$^\dagger$+Barlow Twins) achieves 97.4\% on STL10, and ESSL(DINO+Barlow Twins) achieves 92.9\% on CIFAR10 and 81.6\% on CIFAR100. These transfer learning results comprehensively demonstrate the strong generalizability achieved by ESSL on large-scale benchmark datasets.

\begin{table*}
    \setlength{\tabcolsep}{21.pt}
    \renewcommand\arraystretch{1.3}
    \caption{Linear probing results on the large-scale benchmark, i.e., ImageNet.}
    \label{tab:main_exp_imagenet}
    \centering
    \adjustbox{max width=\linewidth}{
    \begin{tabular}{lllllll}
        \toprule
        Methods                    & Architecture   & Param & Epoch      & Top-1          & Top-5    \\
        \midrule
        SimCLR                     & ResNet-50(1x)  & 25M   & 1000       & 69.3           & 89.0  \\
        MoCo                       & ResNet-50(1x)  & 25M   & 1000       & 71.1           & -     \\
        CMC                        & ResNet-50(1x)  & 25M   & 1000       & 66.2           & 87.0  \\
        CPC                        & ResNet-50(1x)  & 25M   & 1000       & 63.8           & 85.3  \\
        InfoMin Aug                & ResNet-50(1x)  & 25M   & 1000       & 73.0           & 91.1  \\
        Barlow Twins                & ResNet-50(1x)  & 25M   & 1000       & 73.5           & 91.0  \\
        BYOL                       & ResNet-50(1x)  & 25M   & 1000       & 74.3           & 91.6  \\
        SSL-HSIC                   & ResNet-50(1x)  & 25M   & 1000       & 72.2           & 90.7  \\
        VICReg                     & ResNet-50(1x)  & 25M   & 1000       & 73.2           & 91.1  \\
        SwAV                       & ResNet-50(1x)  & 25M   & 800        & 75.3           & -     \\
        DINO                       & ResNet-50(1x)  & 25M   & 800        & 75.3           & -     \\
        MV-MR                       & ResNet-50(1x)  & 25M   & 1000        & 74.5           & 92.1 \\
        GroCo                       & ResNet-50(1x)  & 25M   & 800       & 73.9           & 91.6   \\
        MAE                        & ViT-B-32       & 86M   & 1600       & 68.6           & -     \\
        CAE                        & ViT-B-32       & 86M   & 1600       & 71.4           & -     \\
        \hline
        ESSL(SimCLR$^\dagger$+Barlow Twins)  & ResNet-50(1x)  & 25M   & 1000       & 74.5           & 91.9  \\
        ESSL(DINO+Barlow Twins)     & ResNet-50(1x)  & 25M   & 800        & \textbf{75.8}  & \textbf{93.0}  \\
        \bottomrule
    \end{tabular}}
\end{table*}

\begin{table}
    \renewcommand\arraystretch{1.3}
    \caption{Transfer learning results on the large-scale benchmark with the standard ResNet-50 architecture.}
    \label{tab:main_exp_transfer}
    \centering
    \adjustbox{max width=\linewidth}{
    \begin{tabular}{llll}
        \toprule
        Methods         & STL10    & CIFAR10  & CIFAR100   \\
        \midrule
        Supervised      & -        & 93.6     & 78.3       \\
        SimCLR          & -        & 90.6     & 71.6       \\
        BYOL            & -        & 91.3     & 78.4       \\
        \hline
        ESSL(SimCLR$^\dagger$+Barlow Twins) & \bf 97.4 & 91.7     & 78.0       \\
        ESSL(DINO+Barlow Twins)   & 97.1     & \bf92.9  & \bf81.6    \\
        \bottomrule
    \end{tabular}}
\end{table}

\begin{table*}[t]
    \setlength{\tabcolsep}{8pt}
    \renewcommand\arraystretch{1.08}
    \centering
    \caption{Results on conventional benchmark datasets with different projection head numbers.}
    \begin{tabular}{llllll}
            \toprule
            \multirow{2}*{\textbf{Pre-training}}   & Head  & \multicolumn{4}{c}{\textbf{Evaluation}}                           \\\cline{3-6}
                                                   & Num   & STL10         & CIFAR10       & CIFAR100      & Tiny-ImageNet     \\
            \hline
            \multirow{2}*{STL10}                   & 1     & \textbf{86.0} & \textbf{73.3} & 41.4          & \textbf{29.7}     \\   
                                                   & 2     & 83.6          & 72.4          & \textbf{41.7} & 27.5              \\     
            \hline
            \multirow{2}*{CIFAR10}                 & 1     & \textbf{73.4} & \textbf{83.5} & \textbf{44.0} & \textbf{22.4}     \\   
                                                   & 2     & 62.8          & 67.3          & 36.9          & 16.3              \\      
            \hline
            \multirow{2}*{CIFAR100}                & 1     & \textbf{68.3} & \textbf{73.0} & \textbf{49.9} & \textbf{21.7}     \\   
                                                   & 2     & 63.5          & 67.4          & 35.9          & 15.7              \\    
            \hline
            \multirow{2}*{Tiny-ImageNet}           & 1     & \textbf{76.4} & \textbf{72.8} & \textbf{43.1} & \textbf{35.1}     \\   
                                                   & 2     & 69.2          & 63.7          & 33.5          & 21.3              \\    
            \bottomrule
    \end{tabular}
    \label{tab:ablation_1_2_p}
\end{table*}

\subsection{Generalizability and Discriminability of Introducing Different Loss Functions on Different Datasets}\label{sec:expgen_dis}
We conducted experiments to investigate the impact of introducing two heterogeneous loss functions on generalizability and discriminability. The experimental settings are detailed in \textbf{Appendix \ref{app:impdetails}}, and the results are presented in Figure \ref{fig:gd}. Implementing DINO or SwAV as the generalizability model and Barlow Twins as the discriminability model—i.e., ESSL(DINO+Barlow Twins) and ESSL(SwAV+Barlow Twins)—yields higher performance in both generalizability and discriminability compared to all baselines. Additionally, ESSL(SimCLR$^\dagger$+Barlow Twins) achieves higher generalizability than the baselines. These results indicate that ESSL effectively leverages the strengths of both models, achieving superior overall performance.

Specifically, to separately explore the impacts of loss functions on the generalizability and discriminability of self-supervised learning models, we examine two representative self-supervised loss functions, InfoNCE \cite{2018RepresentationOord} and Barlow Twins \cite{2021Barlow}, in an incremental manner. The comparisons are conducted under identical settings, with implementation details provided in \textbf{Appendix \ref{app:hyper-param}}. The variants are named according to the applied loss functions, e.g., SimCLR$^\dagger$ and Barlow Twins. We pre-train these variants on four benchmark datasets. As shown in Figure \ref{fig:inbt}, when both pre-training and linear probing are performed on the same dataset, models pre-trained with Barlow Twins \cite{2021Barlow} loss outperform those pre-trained with InfoNCE \cite{2018RepresentationOord} loss. However, when pre-training and linear probing are conducted on different datasets, models pre-trained with InfoNCE loss outperform those pre-trained with Barlow Twins loss. Thus, using Barlow Twins \cite{2021Barlow} as the loss function enhances discriminability, while InfoNCE \cite{2018RepresentationOord} as the loss function improves generalizability.


\subsection{Ablation: Influence of the Number of Projection Head}\label{sec:expprojnum}
When we combine the advantages of the two loss functions, i.e., $\mathcal{L}_{GEN}$ and $\mathcal{L}_{DIS}$, we intend to avoid the performance degeneration incurred by the variance of the number of projection heads. To this end, we conduct experiments to choose the best number of projection heads. We follow the settings in \textbf{Section \ref{sec:expsmall}} and compare the generalizability and discriminability of two pre-trained models, where the model, with only one projection head, shares the projection head with two loss functions. The model, with two projection heads, back-propagates the gradients of the two loss functions to the backbone network, respectively. The experimental results are reported in Table \ref{tab:ablation_1_2_p}, where we can observe a significant performance gap between the model with one projection head and the model with two projection heads. Following the consistent experimental settings, we implement the one projection head structure as the network structures of our proposed ESSL and the corresponding ablation models.

\begin{table*}[t]
    \setlength{\tabcolsep}{8pt}
    \renewcommand\arraystretch{1.3}
    \centering
    \caption{Results on conventional benchmark datasets with the proposed ESSL and various ablation models, where SimCLR$^\dagger$ denotes the model pre-trained only with $\mathcal{L}_{GEN}$. Barlow Twins denotes the model pre-trained only with $\mathcal{L}_{DIS}$. ESSL w/ Ensemble denotes a model pre-trained with the simple addition of $\mathcal{L}_{GEN}$ and $\mathcal{L}_{DIS}$. ESSL w/ EGT denotes a model pre-trained with the addition of $\mathcal{L}_{GEN}$ and $\mathcal{L}_{DIS}$ guided by the static EGT saddle point. ESSL w/ EGT+RL denotes a model pre-trained with the addition of $\mathcal{L}_{GEN}$ and $\mathcal{L}_{DIS}$ guided by RL-based dynamic trade-off points.}
    \begin{tabular}{llcccc}
            \toprule
            \multirow{2}*{\textbf{Pre-training}}   & \multirow{2}*{Model}  & \multicolumn{4}{c}{\textbf{Evaluation}}    \\\cline{3-6}
                                                &                   & STL10   & CIFAR10  & CIFAR100  & Tiny-ImageNet \\
            \hline
            \multirow{5}*{STL10}                & SimCLR$^\dagger$     & 84.1    & \bf75.2  & 44.7      & 33.1    \\ 
                                                & Barlow Twins         & 85.5    & 70.0     & 41.5      & 29.8    \\
                                                & ESSL w/ Ensemble     & 84.0    & 74.8     & 44.3      & 27.5    \\\cline{3-6}   
                                                & ESSL w/ EGT          & 86.0    & 73.3     & 41.4      & 29.7    \\
                                                & ESSL w/ EGT+RL       & \bf86.1 & 73.4     & \bf45.4   & \bf33.7 \\
            \hline
            \multirow{5}*{CIFAR10}              & SimCLR$^\dagger$     & 64.8    & 74.4     & \bf45.8   & \bf26.1 \\
                                                & Barlow Twins         & 73.1    & 83.0     & 43.8      & 23.1    \\
                                                & ESSL w/ Ensemble     & 72.5    & 79.0     & 41.9      & 22.1    \\\cline{3-6}   
                                                & ESSL w/ EGT          & 73.4    & 83.5     & 44.0      & 22.4    \\  
                                                & ESSL w/ EGT+RL       & \bf73.8 & \bf84.3  & 44.1      & 22.8    \\
            \hline
            \multirow{5}*{CIFAR100}             & SimCLR$^\dagger$     & 69.2    & 74.8     & 48.5      & \bf25.5 \\
                                                & Barlow Twins         & 69.5    & 73.8     & \bf51.1   & 23.7    \\
                                                & ESSL w/ Ensemble     & 63.5    & 71.4     & 47.1      & 25.2    \\\cline{3-6}   
                                                & ESSL w/ EGT          & 68.3    & 73.0     & 49.9      & 21.7    \\  
                                                & ESSL w/ EGT+RL       & \bf70.0 & \bf75.1  & \bf51.1   & 23.2    \\
            \hline
            \multirow{5}*{Tiny-ImageNet}        & SimCLR$^\dagger$     & 75.2    & 70.6     & 40.5      & 32.6    \\
                                                & Barlow Twins         & 76.0    & 70.0     & 39.9      & 32.7    \\
                                                & ESSL w/ Ensemble     & 74.1    & 70.0     & 40.2      & 30.1    \\\cline{3-6}   
                                                & ESSL w/ EGT          & 76.4    & 72.8     & \bf43.1   & \bf35.1 \\    
                                                & ESSL w/ EGT+RL       & \bf76.8 & \bf73.4  & 40.5      & 32.0    \\
            \bottomrule
    \end{tabular}
    \label{tab:ablations}
\end{table*}

\begin{figure}
    \centering
    \vspace*{0pt}
    \includegraphics[width=0.5\textwidth]{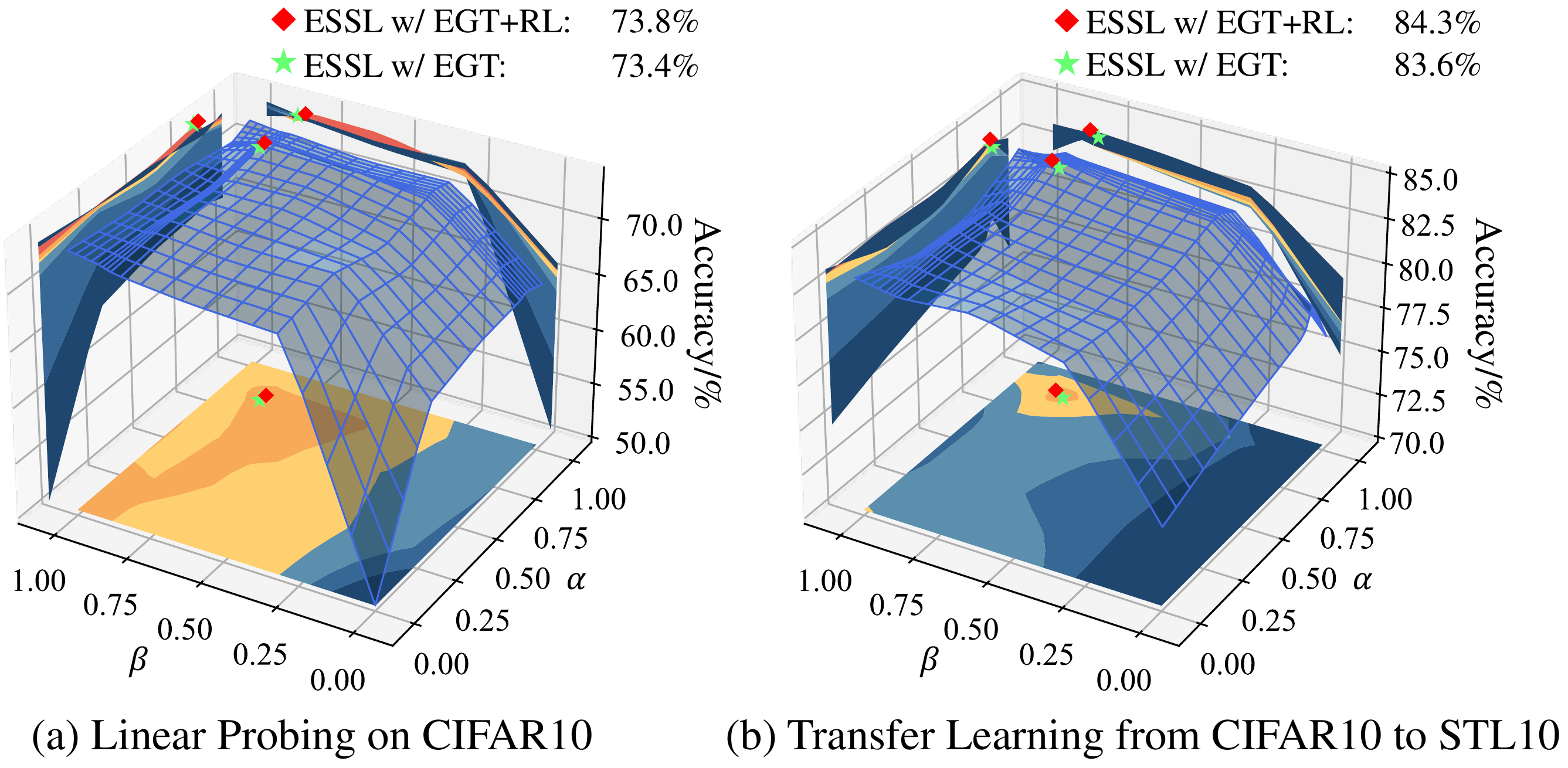}
    \caption{Generalizability and discriminability evaluation results with a wider range of hyper-parameter values for $\alpha$ and $\beta$. The models are pre-trained on CIFAR10. (a) shows the discriminability results on CIFAR10. (b) shows the generalizability results on STL10. The color blocks on the coordinate axis plane represent the accuracy of downstream tasks projected onto the coordinate axis. The effectiveness of the projection decreases as the color shifts toward deep blue, while it increases as the color shifts toward orange.}
    \label{fig:3dplot}
\end{figure}

\subsection{Ablation: Influence of the EGT Module}\label{sec:expegt}
Does the conducted EGT module really improve ESSL's generalizability and discriminability? To answer this question, we conduct an ablation study to analyze the influence of the proposed EGT analysis upon the generalizability and discriminability of the ESSL model. Specifically, we pre-train ESSL models with $\mathcal{L}_{GEN}$ (e.g. SimCLR$^\dagger$, $\alpha$ = 1 and $\beta$ = 0), $\mathcal{L}_{DIS}$ (e.g. Barlow Twins, $\alpha$ = 0 and $\beta$ = 1), ensembled loss (e.g. ESSL w/ Ensemble, $\alpha$ = 1 and $\beta$ = 1) and EGT-guided loss (e.g. ESSL w/ EGT, $\alpha=0.85$ and $\beta=0.87$), then evaluate the generalizability and discriminability of the models on STL10, CIFAR10, CIFAR100, and Tiny-ImageNet. The results are shown in Table \ref{tab:ablations}.
By observing the performance gap between ESSL w/ EGT and ESSL w/ Ensemble, we derive that when pre-training ESSL with the guidance of the EGT analysis, the generalizability and discriminability of ESSL are widely improved. We argue that when chasing better generalizability and discriminability, the position of the trade-off point in the EGT analysis is pivotal, and simply changing the position of the trade-off point or changing the direction of the trade-off point from the origin may incur undesired degeneration on the generalizability and discriminability of ESSL.

To further validate the effectiveness of the EGT module, we conduct comprehensive experiments including a wider range of hyper-parameter values for $\alpha$ and $\beta$, i.e., $\alpha$ and $\beta$ are selected from 0 to 1. Specifically, we pre-train the SSL model with different combinations of hyper-parameters $\alpha$ and $\beta$ on the CIFAR10 dataset and evaluate their discriminability and generalizability on CIFAR10 and STL10, respectively. 
The results are shown in Figure \ref{fig:3dplot}, where the green star represents the performance of the ESSL w/ EGT model, and the red diamond represents the performance of ESSL w/ EGT+RL. We discover the following empirical patterns: 1) as the simpler model with values of alpha and beta approach those of EGT, the model's generalization and discrimination capabilities gradually improve; 2) the model guided by EGT analysis achieves relatively high values in both generalization and discrimination performance, which further indicates that EGT has a significant guiding value.
 
\subsection{The Improvement Derived by Introducing Reinforcement Learning upon the EGT Analysis} \label{sec:expsaddle}

We conduct visualization experiments to further understand the improvement derived by introducing the RL-based optimization upon ESSL. In detail, the proposed EGT analysis provides a significant initial trade-off point of the generalizability and discriminability for ESSL, and RL further empowers ESSL to dynamically adapt the trade-off point to the training domain, i.e., the training dataset, which is achieved by leveraging the Markov Decision Processes.

\begin{figure}
    \centering
    \includegraphics[width=1.0\linewidth]{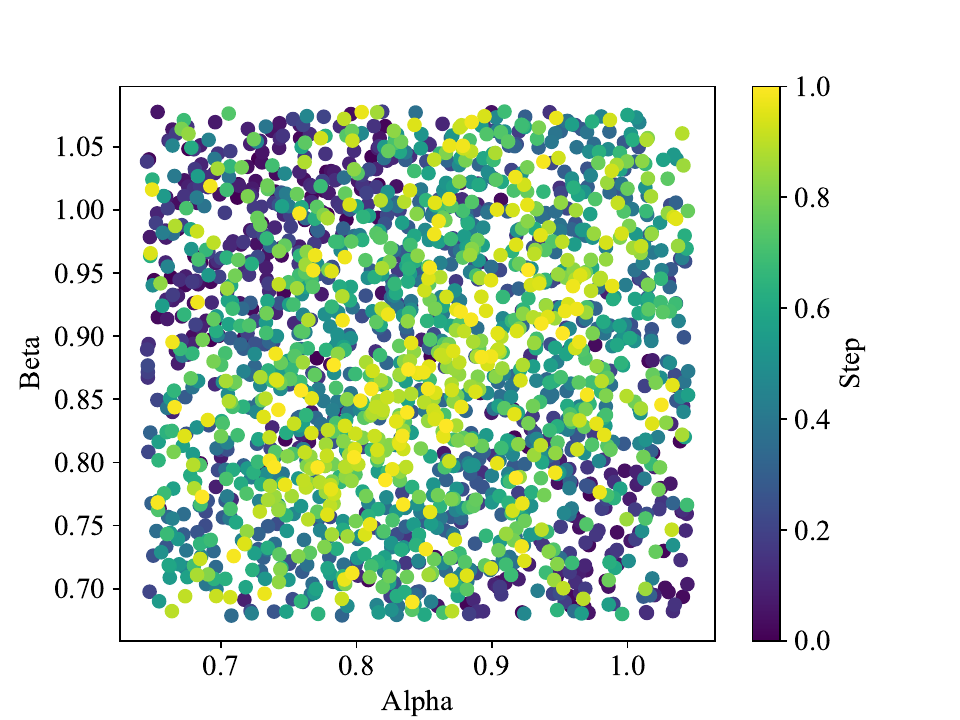}
    \caption{Trade-off points at different stages of pre-training. As the pre-training processes, the points produced by ESSL gradually change from purple to yellow. Note that Alpha and Beta in the figure represent $\alpha$ and $\beta$ in Equation \ref{eq:essl}, respectively.}
    \label{fig:rl_point}
\end{figure}

For the ease of observation of the process and analysis towards the experimental results, we sample 2,000 points uniformly from all 194,560 points records of the hyper-parameters, i.e., $\alpha$ and $\beta$, derived by the RL module of ESSL during the pre-training under the same settings as \textbf{Section \ref{sec:expsmall}} on STL10. The visualization is shown in Figure \ref{fig:rl_point}, where the horizontal axis is $\alpha$, i.e., \textit{Alpha}, and the vertical axis is $\beta$, i.e., \textit{Beta}, and the points produced in the early pre-training stage are inclined to purple. The points produced in the late pre-training stage are inclined to yellow. Specifically, the early pre-training stage denotes that the network has just been initialized or pre-trained with a few epochs, such as 50 or 100. In contrast, the late pre-training stage denotes that the network has been pre-trained after several epochs, such as 900 or 1000.

From the visualization results, we can observe that the points in the early stage of pre-training are distributed in various parts of the area. However, as pre-training is processed, the points come closer to the saddle point derived by the EGT analysis. At the end of pre-training, RL-based ESSL performs dynamic modifications based on the fixed saddle point derived by the EGT analysis, thereby achieving the optimal trade-off between generalizability and discriminability in a dynamic manner. We argue that on the one hand, the EGT analysis can successfully guide ESSL to chase the trade-off between generalizability and discriminability, and the guidance of the fixed saddle point, derived by the EGT analysis, is solid; on the other hand, introducing the RL-based optimization can indeed dynamically update the trade-off point within leveraging the guidance of the EGT analysis in a self-paced manner. As shown in Table \ref{tab:ablations} and Figure \ref{fig:3dplot}, the empirical proofs demonstrate that the generalizability and discriminability of the ultimate ESSL model, i.e., ESSL w/ EGT+RL, widely outperform that of the ablation mode, i.e., ESSL w/ EGT. Note that, due to the dynamic updating mechanism of RL towards $\alpha$ and $\beta$ during the phase of convergence, $\alpha$ and $\beta$, derived by ESSL w/ EGT+RL, are fluctuating in a limited range surrounding the saddle point of EGT analysis, i.e., ESSL w/ EGT. Thus, we adopt the mean values of $\alpha$ and $\beta$ derived by ESSL w/ EGT+RL on convergence to accomplish the visualization.

\begin{figure*}
	\centering
        \includegraphics[width=1.0\textwidth]{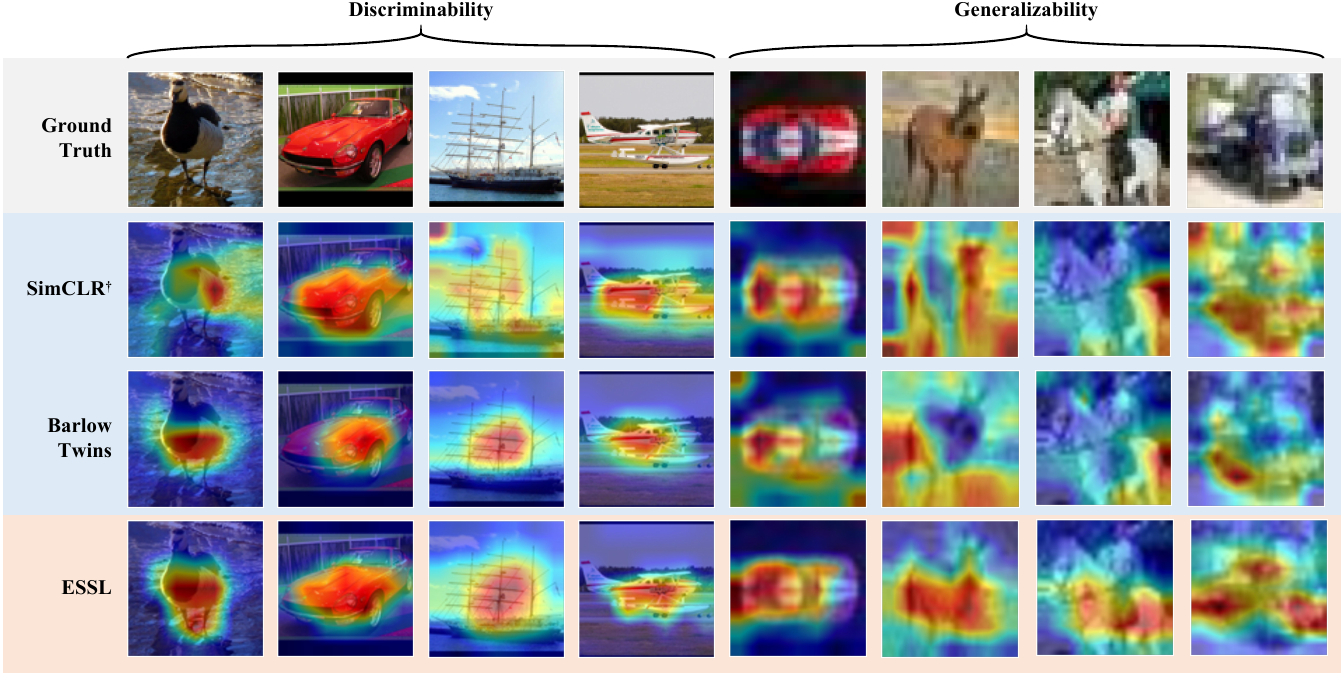}
	\caption{The Grad-CAM \cite{DBLP:journals/ijcv/SelvarajuCDVPB20} visualization comparisons of SimCLR$^\dagger$, Barlow Twins, and ESSL. The comparisons are conducted on STL10 and CIFAR10.}
	\label{fig:visualexplanation}
	\vspace{-0.2cm}
\end{figure*}

\subsection{The Visual Explanation of the Derived Trade-off Mechanisms} \label{sec:visualexplanation}
It is significant and intuitive to explain our disclosed mutual-exclusion between generalizability and discriminability and the proposed trade-off mechanisms by introducing the visualization comparisons. For performing visualization comparisons, we select SimCLR$^\dagger$ (as the generalizability model), Barlow Twins (as the discriminability model), and our proposed ESSL as the candidates. The generalizability comparisons are conducted by training models on STL10 and testing models on CIFAR10, respectively. The discriminability comparisons are conducted by training and testing models on STL10. The visualization results are demonstrated in Figure \ref{fig:visualexplanation}, which are achieved by adopting Grad-CAM \cite{DBLP:journals/ijcv/SelvarajuCDVPB20}.

In Figure \ref{fig:visualexplanation}, we disclose the following empirical patterns: 1) SimCLR$^\dagger$, as the generalizability model, consistently captures the \textit{global} semantic information, while falls short in sufficiently exploring the detailed local semantic information; 2) Barlow Twins, as the discriminability model, can generally focus on the detailed and discriminative \textit{local} semantic information, but the global semantic information is not comprehensively captured; 3) by contrast, the proposed ESSL is able to effectively capture the global semantic information without overly discarding the discriminative local semantic information.

Empirically, from the results of discriminability comparisons, i.e., the leftmost four columns of images, we observe that Barlow Twins emphasizes the discriminative local semantic information, while SimCLR$^\dagger$ learns the global semantic information, resulting in the representations learned by SimCLR$^\dagger$ contain excessive \textit{noises}. Thus, within the \textit{in-distribution} scenarios, e.g., the discriminability comparisons, Barlow Twins outperforms SimCLR$^\dagger$ by a wide margin. However, from the results of generalizability comparisons, i.e., the rightmost four columns of images, we observe that Barlow Twins fails to locate the discriminative local semantic information and captures certain non-discriminative local semantic information. The lack of Barlow Twins for exploring the global semantic information degenerates its ability to correct the biased predictions. On the contrary, SimCLR$^\dagger$ captures the sufficient global semantic information, empowering it to debiasing the prediction within the \textit{out-of-distribution} scenarios. Without loss of generality, the visualization evidence further demonstrates the soundness of the mutual-exclusion between generalizability and discriminability.

Our proposed ESSL seeks an optimal trade-off mechanism between generalizability and discriminability. From the visualization comparisons, shown in Figure \ref{fig:visualexplanation}, ESSL jointly leverages the advances of the generalizability model, i.e., SimCLR$^\dagger$, and the discriminability model, i.e., Barlow Twins, such that ESSL can capture the global semantic information without discarding overmuch local semantic information. Accordingly, in the discriminability comparisons, ESSL learns the \textit{multiple} discriminative semantic information instead of the \textit{single} or \textit{several} discriminative semantic information. In the generalizability comparisons, although the noise caused by the out-of-distribution generalization exacerbates ESSL’s acquisition of the semantic information, the aforementioned characteristics can still enable ESSL to perform relatively robust and debiased predictions. In a nutshell, the visualization comparisons consistently demonstrate the optimality of the trade-off mechanisms derived by the proposed ESSL.

\section{Limitations and Broader Impacts}
The optimization potential exists for ESSL's training process to reduce complexity. Accordingly, acquiring the guidance of EGT requires specific representative downstream datasets with annotations, such that performing EGT analysis in an unsupervised or semi-supervised manner is promising scientific research for the community.

\section{Conclusion}
We disclose a mutual-exclusion relationship between the generalizability and discriminability of learned representations by conducting empirical explorations, which undermines the development of state-of-the-art self-supervised methods. To this end, we revisit the SSL paradigm from the EGT perspective and explore achieving a desired trade-off between generalizability and discriminability. Imposing the guidance of the EGT and the RL, we propose a novel SSL method called ESSL. To comprehensively evaluate the representation quality, we provide a novel self-supervised benchmark. The theoretical evidence demonstrates that ESSL tightens the generalization error bound of SSL. The experimental results prove that our method achieves the state-of-the-art performance on multiple benchmarks.

\section*{Data Availability}
The datasets used in this paper are publicly available. Their names and links are as follows:
\begin{enumerate}
  \item STL10 \nolinkurl{https://cs.stanford.edu/\~acoates/stl10/}.
  \item CIFAR10 \nolinkurl{https://www.cs.toronto.edu//~kriz/cifar.html}. 
  \item CIFAR100 \nolinkurl{https://www.cs.toronto.edu//~kriz/cifar.html}.
  \item Tiny-ImageNet \nolinkurl{https://www.kaggle.com/c/tiny-imagenet}.
  \item ImageNet \nolinkurl{https://image-net.org/index.php}.
\end{enumerate}

\section*{Acknowledgments}
The authors would like to thank the anonymous reviewers for their valuable comments. This work is supported by the China Postdoctoral Science Foundation, Grant No. 2024M753356, the Postdoctoral Fellowship Program, Grant No. GZC20232812, the Postdoctoral Fellowship Program, Grant No. GZB20230790, the Fundamental Research Program, China, Grant No. JCKY2 022130C020, 2023 Special Research Assistant Grant Project of the Chinese Academy of Sciences.

%
%

\bibliographystyle{spbasic}      
\bibliography{ref}   


\clearpage
\appendix

\section{Nomenclature}
\label{app:nomenclature}
For ease of understanding the detailed usage of abbreviations and symbols, we provide the nomenclature in Table \ref{tab:Nomenclature of preliminary and method} and Table \ref{tab:Nomenclature of theoretical analyses}.
\begin{table*}
\centering
\caption{Nomenclature of preliminary and method}
\setlength{\tabcolsep}{30pt}
\renewcommand\arraystretch{1.08}
\label{tab:Nomenclature of preliminary and method}
\begin{tabular}{ll}
    \toprule
    \textbf{Nomenclature} &\\\hline
    \textbf{Variables} &\\
    $\mathcal{D}$ & dataset \\
    $M$ & model \\
    $M_g$ & the generalizability model \\
    $M_d$ & the discriminability model \\
    $M_{ens}$ & the ensemble model of $M_g$ and $M_d$ \\
    $G$ & the gain of generalizability by $M$ \\
    $G_1$ & the gain of generalizability by $M_g$\\
    $G_2$ & the gain of generalizability by $M_d$\\
    $D$ & the gain if discriminability by $M$ \\
    $D_1$ & the gain of discriminability by $M_g$\\
    $D_2$ & the gain of discriminability by $M_d$\\
    $N_1$ & the negative impact of $M_g$ on the discriminability of $M_d$\\
    $N_2$ & the negative impact of $M_d$ on the generalizability of $M_g$\\
    $S_G$ & the strategy spaces of the generalizability model\\
    $S_D$ & the strategy spaces of the discriminability model\\
    $A$ & "Adopted" strategy\\
    $U$ & "Unadopted" strategy\\
    $\bf{H}$ & the income matrix of the generalizability model \\
    $\bf{K}$ & the income matrix of the discriminability model \\
    $U_{G1}$ & the expected utility when the generalizability model selects the strategy A\\
    $\overline{U_G}$ &  the corresponding average expected utility of the generalizability models \\
    $F(\cdot)$ & replicator dynamic equation \\
    $\mathcal{L}_{InfoNCE}$ & InfoNCE loss\\
    $\mathcal{L}_{Barlow Twins}$ & Barlow Twins loss\\
    $\mathcal{L}_{ESSL}$ & ESSL loss\\
    $s_t$ & state in step $t$ of agent\\
    $a_t$ & action in step $t$ that agent select\\
    $r_t$ & reward in step $t$ that agent receives\\\hline
    \textbf{Parameters} & \\
    $\omega_1$ & the preference for generalizability\\
    $\omega_2$ & the preference for discriminability\\
    $x$ & the proportion of discriminability models\\
    $y$ & the proportion of generalizability models\\
    $\xi$ & hyper-parameter in the reward function\\ 
    $V$ & hyper-parameter of action  \\
    $a_t^\alpha$ & the hyper-parameter controlling generalizability at time step $t$\\
    $b_t^\beta$ & the hyper-parameter controlling discriminability at time step $t$\\\hline
    \textbf{Acronyms}  & \\
    SSL & self-supervised learning\\
    M-M & masked language/image modeling \\
    EGT & evolutionary game theory\\
    ESSL & evolutionary game-guided self-supervised learning \\
    RL & reinforcement learning \\
    SS-CL & self-supervised contrastive learning\\
    AGI & artificial general intelligence \\
    CV & computer vision\\
    NLP & natural language processing \\
    PPO & Proximal Policy Optimization\\
    SL & supervised learning \\
    ESS & evolutionary stable strategy\\
    RD & replicator dynamic\\
    $\bf{PT}$ & pre-trained dataset \\
    $S10$ & STL10\\
    $C10$ & CIFAR10\\
    $C100$ & CIFAR100\\
    $Tiny$ &  Tiny-ImageNet \\
    \bottomrule
\end{tabular}
\end{table*}

\begin{table*}
\centering
\caption{Nomenclature of theoretical analyses}
\setlength{\tabcolsep}{24pt}
\renewcommand\arraystretch{1.08}
\label{tab:Nomenclature of theoretical analyses}
\begin{tabular}{ll}
    \toprule
    \textbf{Nomenclature} &\\\hline
    \textbf{Variables} &\\
    $\mathcal{L}_{softmax}$ & softmax loss \\
    $f$ & the fixed encoder pre-trained by self-supervised methods \\
    $W$ & linear probing classifier \\
    $T$ & the label on the downstream tasks \\
    $\mathcal{L}_{GE}^{f}$ & the generalization error for encoder $f$ \\[.3ex]
    $\mathcal{L}_{InfoNCE}^f$ & the conventional contrastive learning objective for encoder $f$ \\[.3ex]
    $\mathcal{L}_{Barlow Twins}^f$ & the Barlow Twins loss for encoder $f$ \\[.3ex]
    $N^S$ & the number of training samples \\
    $N^{batch}$ & the size of a mini-batch \\
    $\mathcal{R}_H(\beta)$ & the Radermacher complexity \\
    $J$ & Jacobian matrix of the RD equations \\
    $\mathcal{H}$ & a restricted function hypothesis space \\
    $X$ & training data \\
    $d$ & the dimensionally of the representations learned by $f$ \\
    $N^{neg}$ & the size of negative pairs \\\hline
    \textbf{Parameters} & \\
    $\alpha$ & the hyper-parameter to denote the weight of $\mathcal{L}_{InfoNCE}^f$ \\
    $\beta$ & the hyper-parameter to denote the weight of $\mathcal{L}_{Barlow Twins}^f$ \\
    $\delta$ & the probability hyper-parameter to hold the upper bound of generalization error \\
    \bottomrule
\end{tabular}
\end{table*}

\section{Datasets}
We evaluate our method on five datasets, STL10 \cite{stl10}, CIFAR10 \cite{tinyimagenetcifar10100}, CIFAR100 \cite{tinyimagenetcifar10100}, Tiny-ImageNet \cite{tinyimagenetcifar10100} and ImageNet \cite{2009Feifei}. The details of each dataset are shown in Table \ref{tab:datasets}.

\begin{table*}[t]
    \setlength{\tabcolsep}{13pt}
    \renewcommand\arraystretch{1.08}
    \caption{Characteristics of image datasets used.}
    \label{tab:datasets}
    \centering
    \begin{threeparttable}
    \begin{tabular}{lrrrrrr}
        \toprule
        Dataset            & Classes & OTE \tnote{1} & TE \tnote{2} & VE \tnote{3} & TE \tnote{4} & Accuracy Measure \\
        \midrule
        STL10              & 10      & 113000        & 100000       & 5000         & 8000         & Top-1 accuracy   \\
        CIFAR10            & 10      & 50000         & 45000        & 5000         & 10000        & Top-1 accuracy   \\
        CIFAR100           & 100     & 50000         & 44933        & 5067         & 10000        & Top-1 accuracy   \\
        Tiny-ImageNet  & 200     & 120000        & 100000       & 10000        & 10000        & Top-1 accuracy   \\
        ImageNet           & 1000    & 1281167       & 1271158      & 10009        & 50000        & Top-1 accuracy   \\
        \bottomrule
    \end{tabular}
    \begin{tablenotes}
        \footnotesize
        \item[1] Original train examples
        \item[2] Train Examples
        \item[3] Valid Examples
        \item[4] Test Examples
    \end{tablenotes}      
    \end{threeparttable}
\end{table*}

\section{Implementation Details} \label{app:impdetails}
\subsection{Code Repositories Duplicated}
\label{app:coderepo}
To evaluate each model's generalizability and discriminability, we use open-source code repositories to train baselines and implement our method. For Barlow Twins \cite{2021Barlow}, the repository is \nolinkurl{https://github.com/facebookresearch/barlowtwins}, SimCLR \cite{chen2020simple} is \nolinkurl{https://github.com/sthalles/SimCLR}, MoCo \cite{2020Kaiming} is \nolinkurl{https://github.com/facebookresearch/moco} and BYOL \cite{2020Bootstrap} is \nolinkurl{https://github.com/sthalles/PyTorch-BYOL}. We implement our method based on the settings of Barlow Twins \cite{2021Barlow}. We implement our work on 4 $\times$ NVIDIA Tesla A100 for 72 hours. Since the GPUs are shared with other groups, the training time is just an approximate time rather than an exact one. 

\subsection{The Supervised Learning Results Used for Calculating EGT} \label{app:slr}
For calculating EGT, we use the performance of existing supervised learning on conventional benchmark datasets, which are 99.60\% \cite{stl10sl} on STL10, 99.37\% \cite{cifar10sl} on CIFAR10, 93.51\% \cite{cifar10sl} on CIFAR100 and 72.18\% \cite{tinysl} on Tiny-ImageNet.

\subsection{Hyper-Parameters}\label{app:hyper-param}
In our EGT model, we set the value of $\omega_1$ and $\omega_2$ to 1, which corresponds to our objective that we hope the model can acquire strong generalizability and discriminability in a balanced manner. The details of training hyper-parameters are shown in Table \ref{tab:hyper-param}.
\begin{table}[t]
    \setlength{\tabcolsep}{16pt}
    \renewcommand\arraystretch{1.08}
    \caption{Hyper-Parameters.}
    \label{tab:hyper-param}
    \centering
    \begin{threeparttable}
    \begin{tabular}{lr}
        \toprule
        Hyper-Parameter          & Value\\
        \midrule
        Epoch                    & 500\tnote{1}, 1000\tnote{2}     \\
        Batch Size               & 256                             \\
        Learning Rate Weights    & 0.2                             \\
        Learning Rate Biases     & 0.0048                          \\
        Weight Decay             & 0.0004                          \\
        Projector Size           & 8192-8192-8192                  \\
        lambda                   & 0.0051                          \\
        Temperature              & 0.07                            \\
        N views                  & 2                               \\
        $Z$                      & 1.5                             \\
        $V$                      & 0.5                             \\
        $k$                      & 200                             \\
        \bottomrule
    \end{tabular}
    \begin{tablenotes}
        \footnotesize
        \item[1] For STL10, CIFAR10, CIFAR100, Tiny-ImageNet
        \item[2] For ImageNet
    \end{tablenotes}   
    \end{threeparttable}
\end{table}

\subsection{Experimental Settings of Methods in Figure \ref{fig:motivation}} \label{app:fig1detail}
The experiments on SimCLR$^\dagger$, Barlow Twins, ESSL w/ Ensemble, and ESSL are conducted under identical settings, including data augmentation, hyperparameters, and other relevant settings, with the only variable being the loss function. The experimental settings are based on the official implementation of Barlow Twins (https://github.com/facebookresearch/barlowtwins). SimCLR$^\dagger$ is a variant of SimCLR in which the original loss function is retained, while other settings, such as data augmentation and hyperparameters, are modified to match those of Barlow Twins. This modification ensures a fair comparison by eliminating the influence of extraneous factors, thereby allowing the study to focus exclusively on the impact of different loss functions on generalizability and discriminability. Concretely, the inconsistency between Figure 1 and Table 3 of the manuscript is clarified.

\section{Pseudo-code and Time Complexity}
The details of the training process are shown in Algorithm \ref{alg:essl}. Related symbols can refer to Table \ref{tab:Nomenclature of preliminary and method}.
\label{app:essl_algo}
\begin{algorithm}
	\vskip 0.in
	\begin{algorithmic}
		\STATE {\bfseries Input:} Multi-view dataset ${X}$ with $NV$ views of each sample, minibatch size $n$, and hyper-parameters $\alpha$, $\beta$, $\delta$.
		\STATE {\bf Initialize} The neural network parameters: ${\theta}$ and $\vartheta$ for view-specific encoders $f_{\theta}(\cdot)$ and projection heads $g_{\vartheta}(\cdot)$, $AGENT$ for agent network. $l$ for learning rate of contrastive learning, $l_a$ for learning rate of RL, train agent per $k$ step.
		\REPEAT
		\FOR{$t$-th training iteration (T epochs)}  
                \STATE Iteratively sample minibatch $\widetilde{X} = \left\{ {X_i} \right\}_{i = (t-1)n}^{tn}$. (time cost $t_0$) 
                \STATE $z_t = g_\vartheta(f_\theta(\widetilde{X}))$ (time cost $t_1$) 
                \STATE $\# \ reinforcement \ learning \ training \ step$
                \STATE $s_t=F_s(z_t)$ (time cost $t_2$) 
                \STATE $\alpha_t, \beta_t=a_t=AGENT(s_t)$ (time cost $t_3$) 
                \STATE $r_t= G(\alpha_t, \beta_t, x^\star, y^\star) + \xi\frac{1}{|L_t-\phi L_{t-1}|}$ (time cost $t_4$)
                \STATE Store transition $(s_t, a_t, r_t)$ (time cost $t_5$)
                \IF{$t \% k == 0$}
                    \STATE $AGENT=AGENT+l_a \delta_{AGENT}\mathcal{L}_{PPO}$ (time cost $t_6$ 
                \ENDIF
            \STATE $\# \ contrastive \ learning \ training \ step$ (time cost $t_7$)
            \STATE $\mathcal{L}_{ESSL}=\alpha_t \mathcal{L}_{InfoNCE} + \beta_t \mathcal{L}_{Barlow Twins}$ (time cost $t_8$)
		  \STATE $\theta \leftarrow \theta - \ell\Delta_{\theta} \mathcal{L}_{ESSL}$ (time cost $t_9$)
		\ENDFOR
		\UNTIL $\theta$, $\vartheta$ converge.
	\end{algorithmic}
	\caption{ESSL: Evolutionary game-guided Self-Supervised Learning}
	\label{alg:essl}
\end{algorithm}

Following the pseudo-code above, we further calculate the time complexity of ESSL. 
\begin{equation}
    \begin{aligned}
        T(episode, t)= & t_0+t_1+t_2+t_3+t_4+t_5+t_8+t_9\times T\\
        = & t \times T\\
        = & T.
    \end{aligned}
    \label{equ:timecost}
\end{equation}
Note that ESSL does not change the order of time complexity, which is still a linear complexity related to epoch numbers.

\end{document}